\pdfoutput=1

\documentclass[11pt]{article}

\usepackage[final]{acl}

\usepackage{times}
\usepackage{latexsym}

\usepackage[T1]{fontenc}

\usepackage[utf8]{inputenc}

\usepackage{microtype}

\usepackage{inconsolata}

\usepackage{graphicx}

\usepackage{amsmath}
\usepackage{xcolor}
\usepackage{algorithm}
\usepackage{algpseudocode}
\usepackage{amsmath}
\usepackage{amssymb}
\usepackage{subcaption}
\usepackage{caption}
\usepackage{pifont}
\usepackage{makecell} 
\usepackage{tabularx}
\usepackage{array}
\usepackage{colortbl}
\usepackage{mathtools}   
\usepackage{tablefootnote}

\definecolor{check}{rgb}{0.333, 0.755, 0.545}

\usepackage{newpxtext}
\usepackage{adjustbox}
\usepackage{booktabs}
\usepackage{amssymb}
\usepackage{pifont}
\newcommand{\cmark}{\ding{51}}
\newcommand{\xmark}{\ding{55}}

\makeatletter
\newcommand{\ie}{\emph{i.e.}\@ifnextchar.{\!\@gobble}{}}
\newcommand{\eg}{\emph{e.g.}\@ifnextchar.{\!\@gobble}{}}
\newcommand{\etc}{etc\@ifnextchar.{}{.\@}}
\makeatother
\newcommand{\mycomment}[1]{}

\title{\textbf{GrowOVER}: How Can LLMs Adapt to Growing Real-World Knowledge?}

\author{
\textbf{Dayoon Ko} \quad \textbf{Jinyoung Kim} \quad \textbf{Hahyeon Choi} \quad \textbf{Gunhee Kim} \\
\vspace*{-0.25cm}\\
Seoul National University \\
\texttt{\footnotesize dayoon.ko@vision.snu.ac.kr \quad jiny1623@snu.ac.kr \quad gk0gus0@snu.ac.kr \quad gunhee.kim@snu.ac.kr}\\
\texttt{\footnotesize \href{https://github.com/dayoon-ko/GrowOVER}{https://github.com/dayoon-ko/GrowOVER}}
}
\vspace{-0.25cm}

\begin{document}
\maketitle
\begin{abstract}
In the real world, knowledge is constantly evolving, which can render existing knowledge-based datasets outdated. This unreliability highlights the critical need for continuous updates to ensure both accuracy and relevance in knowledge-intensive tasks. To address this, we propose \textbf{\textproc{GrowOVER-QA}} and \textbf{\textproc{GrowOVER-Dialogue}}, dynamic open-domain QA and dialogue benchmarks that undergo a continuous cycle of updates, keeping pace with the rapid evolution of knowledge. Our research indicates that retrieval-augmented language models (RaLMs) struggle with knowledge that has not been trained on or recently updated. Consequently, we introduce a novel retrieval-interactive language model framework, where the language model evaluates and reflects on its answers for further re-retrieval. Our exhaustive experiments demonstrate that our training-free framework significantly improves upon existing methods, performing comparably to or even surpassing continuously trained language models.
\end{abstract}

\section{Introduction}

In natural language research, many knowledge-intensive tasks have been actively studied, including open-domain QA \citep{kwiatkowski2019natural, joshi2017triviaqa, yang2018hotpotqa}, fact-checking \citep{thorne2018fever}, entity linking \citep{hoffart2011robust}, and open-domain dialogue \citep{dinan2018wizard}, to name a few \citep{petroni2021kilt, levy2017zero}.
Such a knowledge-intensive task is mostly to utilize
world knowledge to generate a proper answer for a given query \citep{lewis2020retrieval}. 
However, the amount of real-world knowledge is often too enormous for models to fully store them in the parameters.
Thus, in most scenarios, a retriever is employed to seek relevant paragraphs or documents in a knowledge bank, such as a vector database, and a generator provides answers based on the retrieved passages or documents \citep{lewis2020retrieval, guu2020retrieval}.
Previous benchmarks \citep{kwiatkowski2019natural, dinan2018wizard, petroni2021kilt} annotate gold answers and the evidence text needed to predict them, and evaluate the retriever using the evidence text and the generator using the gold answers.

\begin{figure}[t]
    \includegraphics[width=0.48\textwidth]{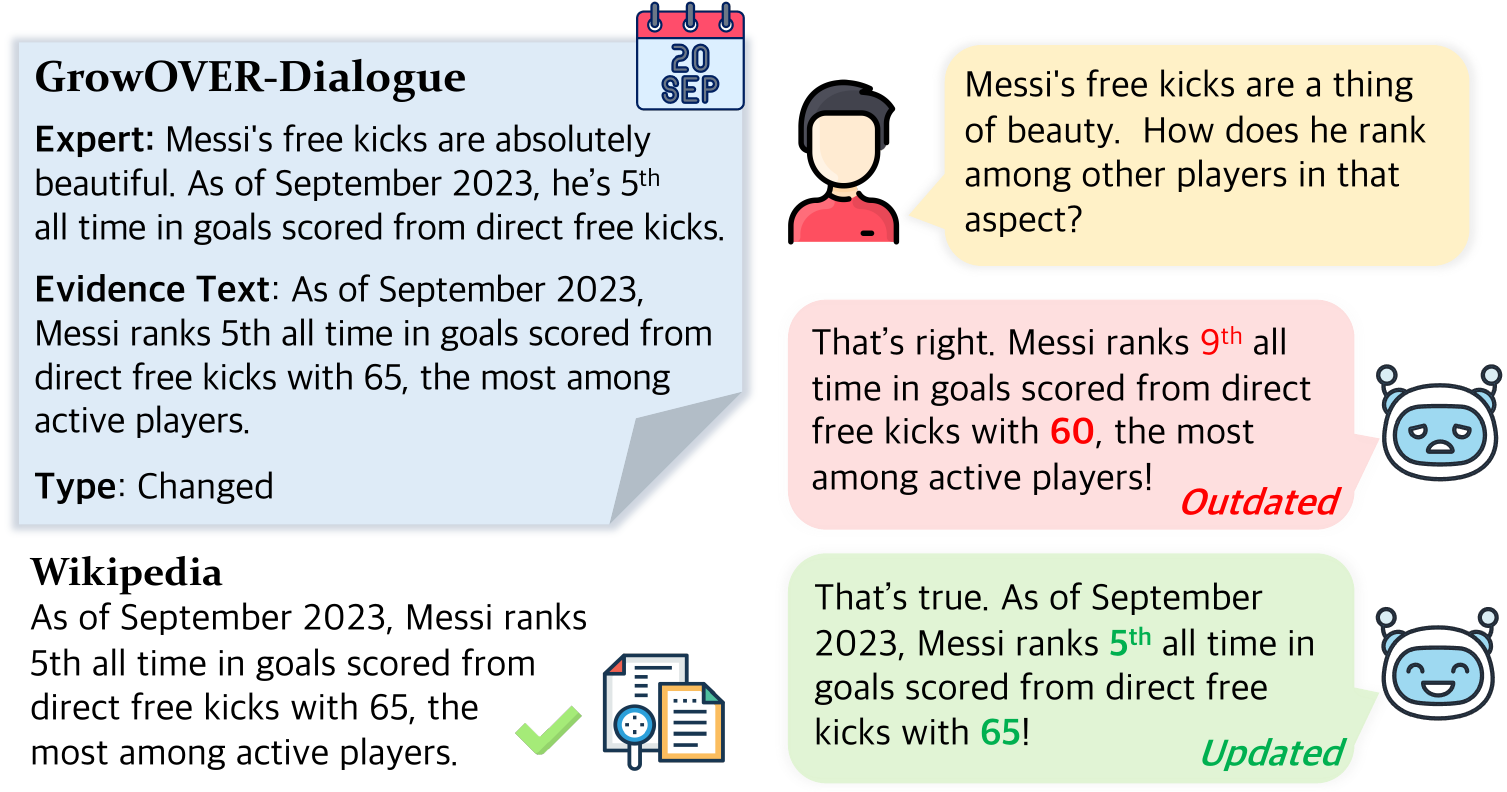}
    \vspace{-0.7cm}
    \caption{\small
    An illustration of \textproc{GrowOVER} benchmarks. \textproc{GrowOVER} is automatically generated and continuously updated. It provides the evidence text to evaluate the retriever and also comprehensively evaluates the generator through an open-domain dialogue task.}
    \label{fig:novelty}
    \vspace{-0.6cm}
\end{figure}

\renewcommand{\arraystretch}{0.8}
\begin{table*}[t]
\centering
\begin{adjustbox}{width=\linewidth}
{\footnotesize
\begin{tabular}{lcccccc}
\toprule
& \textbf{TempLAMA} & \textbf{RealtimeQA} & \textbf{DynamicTempLAMA} & \textbf{TemporalWiki} & \textbf{EvolvingQA} & \textbf{GrowOVER} \\
& \citeyearpar{dhingra2022time} & \citeyearpar{kasai2022realtime} & \citeyearpar{margatina2023dynamic} & \citeyearpar{jang2022temporalwiki} & \citeyearpar{kim2023carpe} & (Ours)
\\\midrule
Label types
    & \textproc{C} 
    & \textproc{C} 
    & \begin{tabular}{ccc}
        \textproc{C}, 
        \textproc{U}, 
        \textproc{N}
    \end{tabular}
    & \begin{tabular}{ccc}
        \textproc{C},
        \textproc{U}
    \end{tabular}
    & \begin{tabular}{ccc}
        \textproc{C},
        \textproc{U},
        \textproc{N}
    \end{tabular}
    & \begin{tabular}{ccc}
        \textproc{C},
        \textproc{U},
        \textproc{N}
    \end{tabular}
    \\\midrule
    Automation
        & \xmark & \xmark & {\color{check}\cmark} & {\color{check}\cmark} & {\color{check}\cmark} & {\color{check}\cmark} \\\midrule
    Maintenance
        & \xmark & \xmark & \xmark & \xmark & \xmark & {\color{check}\cmark} \\\midrule
    Evidence text
        & \xmark & \xmark & \xmark & \xmark & \xmark & {\color{check}\cmark} \\\midrule
    Tasks
    & Cloze query & QA & Cloze query & \xmark & QA & QA \& Dialogue \\\bottomrule
    \end{tabular}}
    \end{adjustbox} 
    \vspace{-0.2cm}
    \caption{\small 
    Comparison of our GrowOVER with existing benchmarks.
    The Label Types display the data types available in each dataset with \textproc{C}hanged, \textproc{U}nchanged, and \textproc{N}ew.
    The automation indicates the feasibility of automatic generation. The Maintenance represents whether the validity of previously generated datasets is verified in the forthcoming time step. The evidence text indicates whether the dataset includes the evidence text. Lastly, the tasks identify the intended tasks for each dataset.
    }
    \label{tab:comparison}
    \vspace{-0.6cm}
\end{table*}

In the real world, new knowledge is constantly being created, and existing knowledge is changing over time, causing annotated benchmarks to become quickly outdated. 
To handle this issue, \citet{kim2023carpe} and \citet{margatina2023dynamic} respectively suggest dynamic QA and cloze query benchmarks, which are automatically generated by comparing two Wikipedia (or Wikidata) snapshots at different times. However, they provide no annotated evidence text for the gold answers, which may make it difficult to evaluate retrievers in open-domain knowledge-intensive tasks. In the retrieval-augmented generation (RAG) framework, it is crucial to accurately measure the performance of each component as well as end-to-end performance. This enables precise identification of error sources and inaccuracies, allowing for less frequent updates to Large Language Models (LLMs). 
Also, the tasks of these benchmarks require the model to provide only direct answers, which mainly consist of entities. In contrast, real-world knowledge can't be structured as simple question-and-answer pairs. Instead, it's more accurately represented as a vast, interconnected knowledge graph. Therefore, there are limitations in evaluating the generator's ability to provide contextually appropriate and informative answers that incorporate relevant background knowledge.

To overcome such limitations, we propose novel open-domain dynamic benchmarks, \textproc{GrowOVER-QA} and \textproc{GrowOVER-Dialogue} (\textbf{Grow}ing \textbf{O}pen-domain knowledge benchmarks for retrie\textbf{V}al-augmented gen\textbf{ER}ation). As illustrated in Figure~\ref{fig:novelty}, \textproc{GrowOVER} provides the evidence text along with the gold answers, which can be used to evaluate the retriever. In addition, we utilize the evidence text to verify the validity of previously generated datasets and maintain valid ones in the succeeding time steps. Consequently, our benchmarks continue to grow from their initial creation as new Wikipedia snapshots continue to come in. 
Furthermore, it introduces a dialogue task to better evaluate the generator. 
The open-domain dialogue task challenges models to adapt to the user's responses and potentially shift topics while still responding accurately. This demands a more sophisticated understanding and application of world knowledge, allowing for a more extensive evaluation. Table~\ref{tab:comparison} presents the comparison of \textproc{GrowOVER} with other benchmarks.

To enable intermittently updated LLMs to cope with the rapidly evolving world, recent research has explored two approaches: retrieval \citep{kasai2022realtime, ram2023context} and continual pretraining \citep{jang2022temporalwiki, kim2023carpe}. The retrieval approach employs a retriever to supply LLMs with new information from an updated database, leveraging their in-context learning capabilities.
The continual pertaining approach updates outdated knowledge within the LLMs' parameters, thereby preventing hallucinations.
However, constantly updating LLMs can be costly and prone to performance degradation, while relying solely on the retriever can be vulnerable.
Therefore, we propose a retrieval-interactive LLM (RiLM). In RiLM, the LLM evaluates its own answers and, if found unreliable, provides feedback to the retriever to locate more relevant documents. The LLM then uses the feedback to generate improved answers. 

Finally, our contributions are as follows.
\begin{enumerate}
    \vspace{-0.18cm}
    \item We introduce \textproc{GrowOVER}, a set of dynamic QA and dialogue benchmarks that evaluate both retrievers and generators by annotating the evidence text and introducing a challenging dialogue task.
    \vspace{-0.18cm}
    \item We propose RiLM, a framework where the LLM evaluates its own answers and provides feedback to the retriever to correct retrievals, thereby regenerating better answers.
    \vspace{-0.18cm}
    \item We empirically demonstrate the effectiveness of our method without requiring additional pre-training of the LLM.
\end{enumerate}

\vspace{-0.05cm}
\section{Related Work}
\vspace{-0.1cm}

\textbf{Temporal sensitivity}.
Temporal misalignment occurs when training and test datasets originate from different time periods.
Past studies \citep{lazaridou2021mind, luu2021time} report poor performance in downstream tasks when making predictions beyond the training period.
Thus, recent research \citep{dhingra2022time, liska2022streamingqa, saxena2021question, jang2022temporalwiki, kim2023carpe} efforts to evaluate 
how LLMs handle time-sensitive information. 
In particular, there have been approaches to utilize a retriever for time-sensitive knowledge. \citet{zhang-choi-2021-situatedqa} and \citet{longpre2021entity} report that, even with an updated evidence corpus, language models trained on previous data struggle to respond to questions in the present. However, \citet{kasai2022realtime} show that LLMs can adjust their generated responses to recently retrieved documents provided by prompting. Still, when failing to retrieve appropriate documents, LLMs may produce outdated answers.
To address this issue, we propose a retrieval-interactive LLM framework that allows the LLM to provide feedback to the retriever to fetch more relevant documents when the answer is less reliable.

\textbf{Retrieval augmented LLM}.
In the initial stages, language models had limited capacity to store a vast amount of factual details.
Hence, prior studies \citep{lewis2020retrieval, guu2020retrieval} introduce RAG, where the generator responds based on the passages provided by the retriever. 
As LLMs grow larger and are pre-trained on huge text corpora, \citet{ram2023context} propose, instead of training LLMs, to combine retrieved content with a query into a prompt for LLMs to generate an answer.
Additionally, \citet{shi2023replug} use an ensemble scheme that provides multiple documents to the LLM, which determines the next token by summing the probability of the next token for each document. 
Recently, studies have focused on when or what to retrieve. For instance, \citet{asai2023self} uses special tokens to decide when to retrieve, and then generates and reflects on the passages and generated answers. Similarly, \citet{jiang2023active} generates a sentence, and if the generated tokens have low probabilities, it retrieves passages using the generated sentence for a long-form generation task. 

\textbf{Continual Knowledge Learning}.
Continual Learning (CL) focuses on training a model on multiple sequential tasks, while retaining knowledge from previously learned tasks and adapting to new ones \cite{chen2018lifelong, he2021analyzing, chen2020recall, 
xu2023parameterefficient, hu2021lora, wang2020kadapter}. 
In the realm of knowledge-intensive tasks, there is an additional imperative for \textit{knowledge revision}. Addressing this, \citet{jang2021towards} introduce the concept of continual knowledge learning (CKL) to manage the dynamic nature of world knowledge. It involves not just retaining previous knowledge but also embracing new information and adapting to updates. These objectives align with the goal of our benchmarks. 

\vspace{-0.2cm}

\vspace{+0.1cm}
\section{The \textproc{GrowOVER} Dataset}\label{sec:dataset}
\vspace{-0.03cm}

\textproc{GrowOVER} comprises two distinct datasets: \textproc{QA} and \textproc{Dialogue}. \textproc{GrowOVER-QA} is designed to evaluate the ability to recall entities, while \textproc{GrowOVER-Dialogue} features user-expert interactions over 3-4 turns to highlight generation capabilities. Each instance from both datasets is annotated with the evidence text and the type: \textproc{Unchanged}, \textproc{Changed}, or \textproc{New}. Our goal is to evaluate the retention of unchanged knowledge, the updating of changed knowledge, and the acquisition of new knowledge, aligning with the objectives of CKL.

\textbf{Article selection}. \textproc{GrowOVER} is based on Wikipedia snapshots\footnote{We download Wikipedia data dumps from \url{https://dumps.wikimedia.org/enwiki/} and use monthly snapshots from 2023-08-20 to 2023-12-20.}, which contain a vast amount of world knowledge. Although there is no limitation to applying our algorithm to generate QA and conversation for any articles, we select the articles linked to \textit{Portal:Current Events} Wikipedia article from January 2023 to December 2023 (About 12K articles).

\mycomment{
\begin{figure}[h]
    \includegraphics[width=0.45\textwidth]{images/overall.pdf}
    \caption{\footnotesize
    The overview of collecting \textproc{GrowOVER}. 
    }
    \label{fig:overall}
    \vspace{-0.4cm}
\end{figure}
}

\begin{figure*}[t!]
    \centering
    \includegraphics[width=0.975\textwidth]{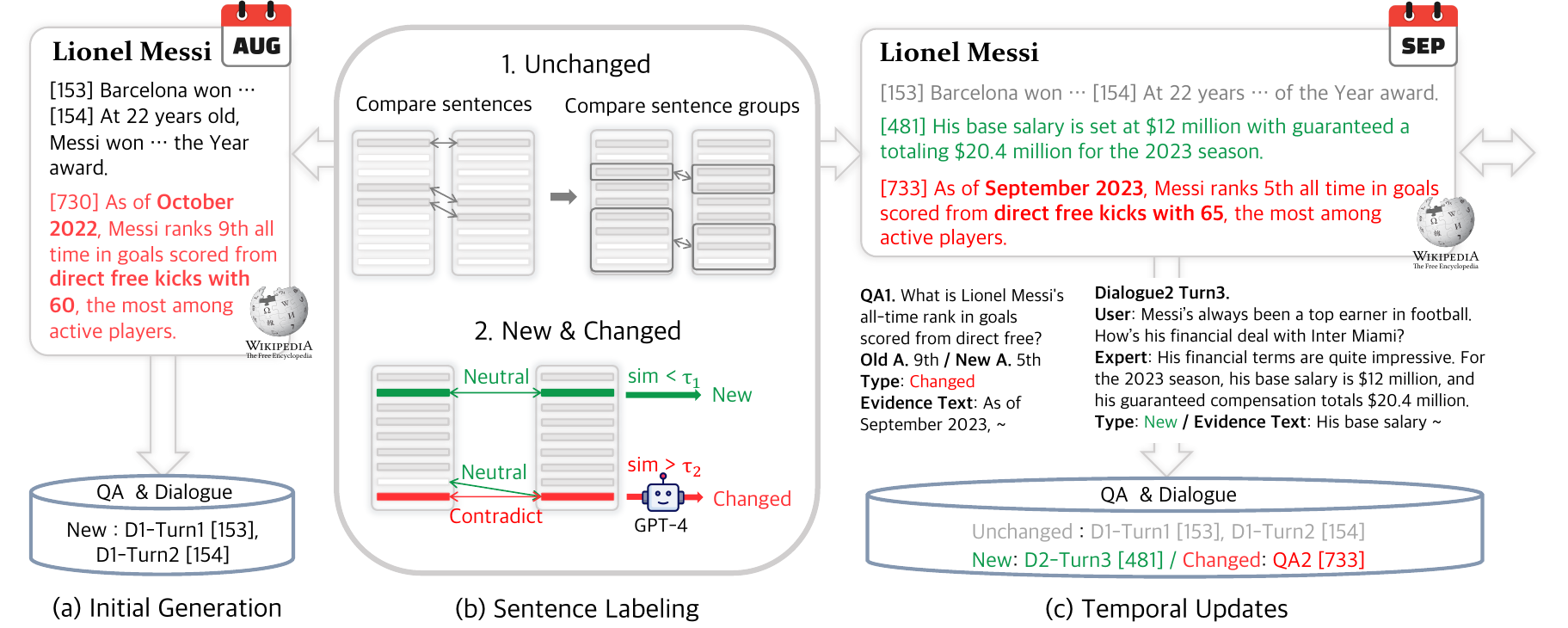}
    \vspace{-0.1cm}
    \caption{\small
    The overview of the dataset generation process. 
    Please refer \S \ref{sec:initial_generation}--\ref{sec:updates} together for detailed explanations.
    }
    \label{fig:labeling}
    \vspace{-0.45cm}
\end{figure*}

\textbf{Overall process}. We create initial QA and dialogue instances using the 2023-08-20 Wikipedia snapshot using GPT-4. 
For each subsequent snapshot, we label each sentence in articles as \textit{unchanged}, \textit{changed}, or \textit{new} by comparing it to the previous month's snapshot. Then, we retain QA and dialogue instances when the evidence text is labeled as \textit{unchanged}, and create new instances from \textit{new} or \textit{changed} sentences.
This process is repeated for each month's new snapshot from September through December. Algorithms and prompt templates for sentence labeling and data generation are detailed in Appendix~\ref{generationAlgs}. Also, the statistics of datasets are provided in Appendix~\ref{dataAnalysis}. 
\vspace{-0.15cm}

\subsection{Initial Generation} \label{sec:initial_generation}

\textbf{\textproc{GrowOVER-QA}}. 
Each article in snapshots in the initial month is split into paragraphs. Then, we select up to four paragraphs, for each of which we prompt GPT-4 to generate QA. The sentences used for generating QA become the evidence text. 
To ensure dataset quality, we experimentally set the criteria for paragraph selection. First, we select paragraphs with less than five sentences and between 300 and 600 characters. If there are more than four satisfying paragraphs, we use K-Means clustering to group them into four clusters and randomly select one paragraph from each cluster. This creates semantically diverse QAs while avoiding too similar QAs.
Afterward, we guide GPT-4 to satisfy the following: i) the question should be directly answered without the context (\eg. no "According to the context), ii) the answers must be short and be entities, and iii) return a bounding box indicating the sentence(s) that include the answer as the evidence text. 

\textbf{\textproc{GrowOVER-Dialogue}}. The paragraph selection process is the same as QA. For each paragraph, we ask GPT-4 to create a dialogue involving user-expert interactions spanning 3-4 turns. We guide it to return the sentence used for generating each turn and then annotate it as the evidence text for each turn. 
\mycomment{
\begin{algorithm}[H]
\caption{Sentence Labeling}
\label{alg:sentence_labeling}
\footnotesize

\textbf{Require:} \texttt{sentences\_old}:= sentences in the old article, \\\texttt{sentences\_new}:= sentences in the new article

\begin{algorithmic}

\State \Comment{Identify \textit{unchanged} sentence pairs}
\For{$\text{s}_\text{old}$ in \texttt{sentences\_old}}
    \For{$\text{s}_\text{new}$ in \texttt{sentences\_new}}
        \If{\texttt{sim}($\text{s}_\text{old}$, $\text{s}_\text{new}$) > thrs(=0.99)}
            \State{($\text{s}_\text{old}$, $\text{s}_\text{new}$) $\leftarrow$ \textit{unchanged}}
        \EndIf
    \EndFor
\EndFor
\\
\Comment{Identify \textit{unchanged} sentence groups}
\For{each group of sentences $(\text{S}_\text{old}, \text{S}_\text{new})$ enclosed by \textit{unchanged} pairs}
    \For{$\text{sub}_\text{old}$ in $\mathcal{P}(\text{S}_\text{old})$}
        \For{$\text{sub}_\text{new}$ in $\mathcal{P}(\text{S}_\text{old})$}
            \If{\texttt{sim} (\texttt{concat}($\text{sub}_\text{old}$), \texttt{concat}($\text{sub}_\text{new}$)) > \\\hspace{5.5em}thrs(=0.99)}
                \State{($\text{sub}_\text{old}$, $\text{sub}_\text{new}$) $\leftarrow$ \textit{unchanged}}
            \EndIf
        \EndFor
    \EndFor
\EndFor
\\
\Comment{NLI for \textit{changed} \& \textit{new}}
\For{each group of sentences $(\text{S}_\text{old}, \text{S}_\text{new})$ enclosed by \textit{unchanged} pairs}
    \For{$\text{s}_\text{new}$ in $\text{S}_\text{new}$}
        \State \texttt{preds} $\leftarrow$ an empty list
        \For{$\text{s}_\text{old}$ in $\text{S}_\text{old}$}
            \State \texttt{preds}.append(\texttt{NLI}.\texttt{classify}($\text{s}_\text{old}$, $\text{s}_\text{new}$))
        \EndFor 
        \If{\text{``entailment''} in \texttt{preds}}
            \State $\text{s}_\text{old}$ $\leftarrow$ the entailed old sentence
            \State ($\text{s}_\text{old}, \text{s}_\text{new}$) $\leftarrow$ \textit{unchanged}
        \ElsIf{\text{``contradiction''} in \texttt{preds}}
            \State $\text{s}_\text{old}$ $\leftarrow$ the contradicted old sentence
            \If{\texttt{sim} ($\text{s}_\text{old}$, $\text{s}_\text{new}$) > $\tau_1$ (=0.6)}
                \If{\texttt{GPT-4}.\texttt{contradict}($\text{s}_\text{old}, \text{s}_\text{new}$)}
                    \State ($\text{s}_\text{old}, \text{s}_\text{new}$) $\leftarrow$ \textit{changed}
                \EndIf
            \EndIf
        \Else \Comment{all elements in \texttt{preds} are ``neutral''}
            \State \texttt{sim\_res} $\leftarrow$ [\texttt{sim}($\text{s}_\text{old}, \text{s}_\text{new}$) for $\text{s}_\text{old}$ in $\text{S}_\text{old}$)]
            \State \texttt{max\_similarity} $\leftarrow$ \texttt{max}(\texttt{sim\_res})
            \If{\texttt{max\_similarity} < $\tau_2$ (=0.7)}
                \State $\text{s}_\text{new}$ $\leftarrow$ \textit{new} 
            \EndIf
        \EndIf
    \EndFor
\EndFor

\end{algorithmic}
\end{algorithm}
\newpage

}
\vspace{-0.15cm}

\subsection{Sentence Labeling}\label{sec:labeling}
\vspace{-0.08cm}
\textbf{Unchanged.} Each time a new snapshot of Wikipedia becomes available, we first identify \textit{unchanged} sentences for each pair of old and new articles. We compute sentence similarity using SimCSE \citep{berant2013semantic} to localize semantically identical sentence pairs, \((\text{s}_\text{old}, \text{s}_\text{new})\). If the similarity score exceeds a threshold of 0.99, we label  \((\text{s}_\text{old}, \text{s}_\text{new})\) as \textit{unchanged}.

\vspace{-0.07cm}
Additionally, we group sentences and compare them to find all unchanged sentences. Wikipedia articles generally maintain the order of sentences even after editing. 
Hence, we group old and new sentences into \(\text{S}_\text{old}\) and \(\text{S}_\text{new}\), respectively, that lie between pairs of previously identified \textit{unchanged} sentences. Then, we compute the similarity between all subsets of \(\text{S}_\text{old}\) and \(\text{S}_\text{new}\). If any subset of \(\text{S}_\text{new}\) matches with a subset of \(\text{S}_\text{old}\) with a similarity score above the threshold, those sentences are labeled as \textit{unchanged}.
For instance, if we have matched \((\text{s}_{\text{old}_1}, \text{s}_{\text{new}_1})\) and \((\text{s}_{\text{old}_4}, \text{s}_{\text{new}_5})\) as \textit{unchanged}, we then compare similarity between all subsets of \(\text{S}_\text{old}=\{\text{s}_{\text{old}_2}, \text{s}_{\text{old}_3}\}\) and \(\text{S}_\text{new}=\{\text{s}_{\text{new}_2}, \text{s}_{\text{new}_3}, \text{s}_{\text{new}_4}\}\). If the similarity score between \(\{\text{s}_{\text{old}_2}\}\) and \(\{\text{s}_{\text{new}_2}, \text{s}_{\text{new}_3}\}\) exceeds the threshold, we label \((\text{s}_{\text{old}_2}, \{\text{s}_{\text{new}_2}, \text{s}_{\text{new}_3}\})\) as \textit{unchanged} as well. After identifying \textit{unchanged} sentences in the groups, we update the groups enclosed by the new pairs of unchanged sentences. In the previous example, \(\text{S}_\text{old}\) becomes \(\{\text{s}_{\text{old}_3}\}\) and \(\text{S}_\text{new}\) becomes \(\{\text{s}_{\text{new}_4}\}\).

\textbf{NLI.} Next, we classify unlabeled sentences in \(\text{S}_\text{old}\) and \(\text{S}_\text{new}\) using a natural language inference (NLI) task with RoBERTa \citep{liu2019roberta} fine-tuned on the MultiNLI dataset \citep{williams-etal-2018-broad}. The NLI task, given a premise and a hypothesis, classifies the hypothesis as entailment, neutral, or contradiction. In this step, we provide the model with each sentence pair \((\text{s}_\text{old}, \text{s}_\text{new})\) where \(\text{s}_\text{old} \in \text{S}_\text{old}\) and \(\text{s}_\text{new} \in \text{S}_\text{new}\) treating \(\text{s}_\text{old}\) as the premise and \(\text{s}_\text{new}\) the hypothesis. If \(\text{s}_\text{new}\) is classified as entailed by any \(\text{s}_\text{old}\), we label that \(\text{s}_\text{new}\) as \textit{unchanged}. 

\textbf{Changed.} Else if \(\text{s}_\text{new}\) is classified as contradiction with any \(\text{s}_\text{old}\), we label the \((\text{s}_\text{old}, \text{s}_\text{new})\) as \textit{changed}. Besides, we check whether their similarity is higher than \(\tau_2\) set to 0.6 since \textit{changed} sentences share some content, not entirely new. After that, we double-check whether \(\text{s}_\text{old}\) and \(\text{s}_\text{new}\) are contradictory with GPT-4 to ensure that \((\text{s}_\text{old}, \text{s}_\text{new})\) is \textit{changed}. 

\textbf{New.} Otherwise, if \(\text{s}_\text{new}\) is classified as neutral, we further check their similarity scores. If the similarity with all \(\text{s}_\text{old} \in \text{S}_\text{old}\) is lower than \(\tau_1\) set to 0.7, we classify that  \(\text{s}_\text{new}\) as \textit{new} since \textit{new} sentence should not have similar counterparts in the old document.

\vspace{-0.1cm}
\subsection{Temporal Updates} 
\vspace{-0.1cm}

\label{sec:updates}
If an article is newly added in the new snapshot, we perform the initial generation as done in~\ref{sec:initial_generation}. Otherwise, based on the results of sentence labeling, we update \textproc{GrowOVER} in two ways: i) maintain or exclude existing QA and dialogue turn instances, and ii) generate new instances. 

\textbf{Maintenance}. Each QA and turn instance is annotated with evidence sentences and their indices within the article. If all evidence sentences \(\text{s}_\text{old}\) are labeled as \textit{unchanged}, we keep the corresponding QA or turn as \textproc{Unchanged} and update the evidence sentences and their indices as of the new snapshot. Otherwise, we exclude the instance since it is not guaranteed as \textproc{Unchanged}. 
For QA, we simply delete the instance from our dataset. But, for dialogue, we only exclude the turn when evaluation instead of the whole dialogue.

\textbf{Generation}. We generate new QA and dialogues with \textit{new} and \textit{changed} sentences. For QA, we find consecutive \textit{new} sentences and split them into multiple groups if more than six sentences. 
For each group, we prompt GPT-4 to generate a \textproc{New} QA instance and annotate the evidence text as done before. 
For \textit{changed} sentences, we provide GPT-4 with both the original and the revised sentences and prompt it to generate a \textproc{Changed} question with contradictory answers based on each sentence and annotate the revised sentences as evidence text. 
For dialogue generation, similarly to the initial generation process, we select informative paragraphs with \textit{changed} or \textit{new} sentences. We then prompt GPT-4 to generate dialogues and annotate the used sentences as the evidence text for each turn. If the evidence text is \textit{changed} or \textit{new}, we label the generated turn as \textproc{Changed} and \textproc{New}, respectively. 
\vspace{-0.1cm}

\begin{figure*}[t!]
    \centering
    \includegraphics[width=0.95\textwidth]{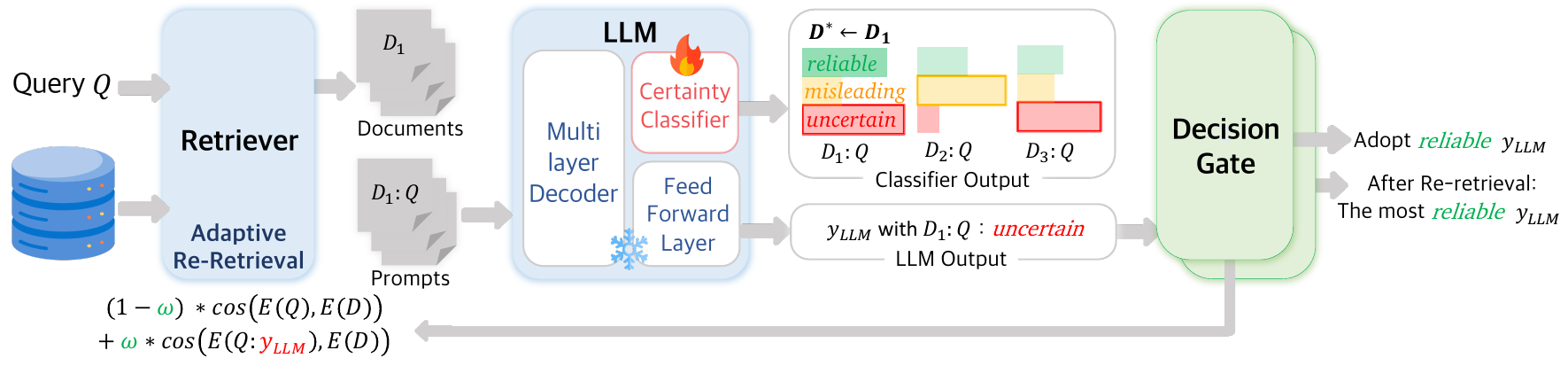}
    \vspace{-0.1cm}
    \caption{\small
    The RiLM framework. Given a query, we retrieve top-$k$ documents and generate $k$ prompts to LLM in parallel. The certainty classifier predicts either \textit{reliable}, \textit{misleading}, or \textit{uncertain} for each prompt. If \textit{reliable}, the Decision Gate adopts the answer. Otherwise, we return back to the retrieval step with LLM's output and the \textit{reliable} probability. In Adaptive Re-retrieval, the retriever reflects this information outputs for better retrieval, based on which the LLM re-generates answers.}
    \label{fig:method}
    \vspace{-0.45cm}
\end{figure*}

\section{Approach}
\vspace{-0.1cm}

If the LLM has been trained on outdated data or has never been trained on new data, it may not be able to answer the questions on new information correctly. However, updating the parameters of such models should be conducted with the greatest caution to avoid potential side effects, such as catastrophic forgetting. Therefore, our approach ensures that neither the LLM nor the retriever is continuously trained with new data.

To adapt LLMs to rapidly evolving world knowledge, we propose RiLM, as shown in Figure~\ref{fig:method}. While freezing both the LLM and the retriever, we introduce the Decision Gate, which decides whether to accept the LLM's answer based on its certainty score (section~\ref{dg}). If the answer is not sufficiently confident, the retrieval-generation process is performed again, termed Adaptive Re-Retrieval (section~\ref{arr}). In this process, the LLM's previous output is fed back to the retriever to fetch documents again, enabling the LLM to generate better answers.

Similar to our approach, \citet{asai2023self} propose using reflection tokens to confirm output relevance, support, or completeness. However, their method requires training LLMs to predict reflection tokens. In contrast, RiLM only involves training the certainty classifier. Additionally, for long-form text generation, \citet{jiang2023active} generate the next sentence and then concatenate it with the question to form a retrieval query. They exclude tokens generated with low probability to avoid interrupting retrieval. Our approach also uses generated answers for re-retrieval but considers all generated tokens only to the extent that the LLM is certain about the answer.

\vspace{-0.1cm}
\subsection{The Decision Gate}\label{dg}
Previous work \citep{ram2023context} has often concatenated top-k retrieved documents \(\{D_1\cdots D_k\}\) with a query \(Q\) into a single prompt to the LLM. However, some irrelevant content may degrade performance. Thus, following \citet{shi2023replug}, we concatenate each document \(D_i\) with the query \(Q\) into a prompt to LLM in parallel.
To select the best answer from these prompts, we add a certainty classifier to the last multi-head attention layer of the LLM.

More specifically, we pass the query and each document through the LLM to obtain the last hidden state vector \(h_{\mathrm{LLM}}(Q, D_i)\). For each hidden state, the certainty classifier predicts either of the following three labels: i) \textit{reliable}: the LLM confidently knows the answer, ii) \textit{misleading}: the LLM knows the answer but could be wrong, or iii) \textit{uncertain}: the LLM does not know the answer exactly.

To train the classifier, we assume that the LLM knows the answer if the data it has been trained on remains unchanged and the retrieval succeeds. Conversely, the LLM might incorrectly know the answer if the data it has been trained on has been updated and the retrieval fails. Lastly, the LLM does not know the answer if it has never been trained on the data and the retrieval fails. Based on this assumption, we train the classifier to predict: i) \textit{reliable}: given \textproc{Unchanged} QA/turn with correct retrieval, ii) \textit{misleading}: given \textproc{Changed} QA/turn with wrong retrieval, iii) \textit{uncertain}: given \textproc{New} QA/turn with wrong retrieval.
We train the classifier separately for QA and dialogue tasks.  
We use 512, 245, and 512 data points of \textproc{Unchanged}, \textproc{Changed}, and \textproc{New} \textproc{GrowOVER-QA}, respectively, and 512, 133, and 512 \textproc{Unchanged}, \textproc{Changed}, and \textproc{New} turns of \textproc{GrowOVER-Dialogue} from September.

We choose the hidden state \(h_{\mathrm{LLM}}(Q, D^{*})\) with which the certainty classifier outputs the highest \textit{reliable} probability:
\begin{equation}\footnotesize
    D^{*} = \operatorname*{arg\,max}_{D\ \in\ \{D_i\}}\ p_{\mathrm{CLF}}\ (\textit{reliable} \mid h_{\mathrm{LLM}}(Q, D)).
\end{equation}
Afterward, LLM generates the answer \(y_{\text{LLM}}\) based on the \(h_{\text{LLM}}(Q, D^{*})\).
If the certainty classifier predicts the label as \textit{reliable} given \(h_{\text{LLM}}(Q, D^{*})\),
the Decision Gate adopts the answer; otherwise, we return to the retrieval step.

\subsection{Adaptive Re-Retrieval}\label{arr}
If the classifier does not predict \textit{reliable}, we re-retrieve documents since all top-k documents are unlikely to be helpful in generating correct answers. Instead of simply retrieving the next set of top-k documents, we propose an Adaptive Re-Retrieval (ARR) method. This method feeds the LLM's answer and certainty value back to the retriever to improve relevance.

In ARR, the retriever relies on the LLM's answer to the extent that it is reliable. 
The \textit{reliable} probability is computed by
\begin{equation}\footnotesize
    \omega = \lambda \ p_{\mathrm{CLF}}\ (\textit{reliable} \mid h_{\mathrm{LLM}}(Q, D^{*})). \label{eqn:prob}
\end{equation}
The hyperparameter \(\lambda\) is set to optimize re-retrieval relevance with the training dataset for the certainty classifier.
Therefore, the relevance score in the ARR is calculated as a weighted sum of two components: cosine similarity between the query and the document, and the similarity between the document and a concatenation of the query with the generated answer \(y_\mathrm{LLM}\). This is formulated as:
\begin{equation}\footnotesize
\begin{split}
    score &= (1 - \omega) \ \mathrm{sim}(\mathbf{E}(Q), \mathbf{E}(D))\\
            &+\ \omega \ \mathrm{sim}(\mathbf{E}([Q:y_{\mathrm{LLM}}]), \mathbf{E}(D)).\label{eqn:score}
\end{split}
\end{equation}
The retriever re-retrieves top-k documents based on Eq.(\ref{eqn:score}). 
We adjust the reflection of the generated answer based on the LLM's predicted \textit{reliable} probability: the less \textit{reliable} an answer is, the less likely it is to be used.
After the re-retrieval, we again choose the last hidden state vector from the document with the highest \textit{reliable} probability among the re-retrieved documents. In the final step, the Decision Gate compares the \textit{reliable} probabilities of the initial and newly generated answers, selecting the one with the higher probability. 

\vspace{-0.05cm}
\section{Experiments}

\vspace{-0.1cm}
\subsection{Experimental Setup}
\vspace{-0.1cm}
\textbf{Baselines}. We use five types of baselines: i) \textbf{Vanilla}: LLM without retrievals, ii) \textbf{Self-RAG}: an adaptive RAG baseline \citep{asai2023self}, \textbf{RaLM}: vanilla LLM with concatenated retrievals \citep{ram2023context}, iii) \textbf{RaLM-CP}: continuously pre-trained LLM \citep{jang2022temporalwiki} with concatenated retrievals, iv) \textbf{RaLM-$D^{*}$}: LLM generates an answer with the classifier's selected document, and v) \textbf{RiLM}. We use top-k (k=3) documents for retrievals. 

\textbf{Database}. Since it requires much computation to use the entire snapshot (6M articles) for the database, we randomly select 100K articles in addition to the 12K articles selected for \textproc{GrowOVER} generation. We first split these articles using the LangChain document loader\footnote{\url{https://python.langchain.com/docs/modules/ data\_connection/document\_loaders/}.}, which semantically segments given documents. Then, we index documents using FAISS \citep{johnson2019billion} following \citet{shi2023replug}.

\textbf{Retriever}. Even though all retrievers can be plugged into our framework, we use SentBERT \citep{reimers2019sentence} since it reduces time cost using Siamese encoders with strong performance. It uses cosine similarity to calculate sentence similarity. We also test Contriever \citep{izacard2021unsupervised} and present the results in Appendix~\ref{addexpResults}. We use questions as retrieval queries for the QA task, and user queries combined with the chat history for the dialogue response task. 

\textbf{LLM}. We use Llama2-7B \citep{touvron2023llama}. We initially pre-train the model with the selected articles in the 2023-08-20 snapshot. For RaLM-CP, we continuously pre-train the model on each new snapshot available every month, using only the selected articles from these snapshots for dataset generation. For the other baselines, we freeze the initially pre-trained model. In the QA task, the prompt consists of the retrieved documents and questions, while in the dialogue task, the prompt concatenates the chat history, retrieved documents, and the user query.


\textbf{Metric}. We use the F1 score to evaluate QA, following \citet{petroni2021kilt} and use the BLEU \citep{papineni2002bleu} score for the dialogue task, follwing \citep{chan2021enhancing}. 

\begin{table}[t!]
\begin{center}
\renewcommand{\arraystretch}{1.2}
\resizebox{\linewidth}{!}{
{\footnotesize
\begin{tabular}{p{4.0cm} >{\centering\arraybackslash}p{0.75cm} >{\centering\arraybackslash}p{0.75cm} >{\centering\arraybackslash}p{0.75cm} >{\centering\arraybackslash}p{0.75cm}}
\Xhline{2\arrayrulewidth}
 & 9 & 10 & 11 & 12\\ 
\hline
\textbf{\textproc{GrowOVER-QA}} \\ 
\hline
    Accuracy & 79.0 & 75.5 & 75.8 & 74.9 \\
    Average F1 (Adopted) & 53.7 & 52.1 & 53.1 & 52.7 \\
    Average F1 (Not-adopted) & 28.2 & 28.5 & 28.8 & 28.6 \\
    Average F1 (ALL) & 43.5 & 42.9 & 42.8 & 42.3\\
\hline
\textbf{\textproc{GrowOVER-Dialogue}} \\ 
\hline
    Accuracy & 59.0 & 58.6 & 58.3 & 58.6 \\
    Average BLEU (Adopted) & 6.03 & 6.11 & 6.17 & 6.13 \\
    Average BLEU (Not-adopted) & 3.44 & 3.42 & 3.44 & 3.45 \\
    Average BLEU (ALL) & 4.68 & 4.69 & 4.72 & 4.70 \\
\Xhline{2\arrayrulewidth}
\end{tabular}
}
}
\vspace{-0.2cm}
\caption{\small Accuracy of the certainty classifier and results of adopted / not-adopted answers on each month.}
\label{tab:clf_sentbert}
\vspace{-0.35cm}
\end{center}
\end{table}

\begin{table}[t!]
\begin{center}
\renewcommand{\arraystretch}{1.2}
\resizebox{\linewidth}{!}{
{\footnotesize
\begin{tabular}{p{4cm} >{\centering\arraybackslash}p{0.8cm} >{\centering\arraybackslash}p{0.8cm} >{\centering\arraybackslash}p{0.8cm} >{\centering\arraybackslash}p{0.8cm}}
\Xhline{2\arrayrulewidth}
 & 9 & 10 & 11 & 12\\  
\hline
\textbf{\textproc{GrowOVER-QA}} \\ 
\hline
    $Q$ & 13.4 & 13.0 & 12.8 & 12.5 \\
    $Q$:$y_\mathrm{LLM}$ & 14.5 & 14.1 & 13.4 & 13.3 \\
    ARR & \textbf{14.6} & \textbf{14.5} & \textbf{13.9} & \textbf{13.6}\\
\hline
\textbf{\textproc{GrowOVER-Dialogue}} \\ 
\hline
    $Q$ & 10.7 & 10.5 & 10.5 & 10.5 \\
    $Q$:$y_\mathrm{LLM}$ & 11.5 & 11.3 & 11.4 & 11.3 \\
    ARR & \textbf{11.7} & \textbf{11.7} & \textbf{11.5} & \textbf{11.5} \\
\Xhline{2\arrayrulewidth}
\end{tabular}
}
}
\vspace{-0.25cm}
\caption{\small Accuracy of ARR on each month.
}
\label{tab:arr_sentbert}
\vspace{-0.65cm}
\end{center}
\end{table}

\vspace{-0.1cm}
\subsection{Experimental Results}
\vspace{-0.1cm}

We report not only the end-to-end performance of the QA and dialogue tasks, but also the performance of the classifier and ARR to demonstrate the effectiveness of each component. 
For all experiments, we exclude the training dataset of the classifier. 

\vspace{-0.1cm}
\subsubsection{Results of the Classifier}
\vspace{-0.1cm}

Experimental results of the classifier are presented in Table~\ref{tab:clf_sentbert}. We evaluate the accuracy of all data points from all months. 
The accuracy is approximately 75 for \textproc{GrowOVER-QA} and 58 for \textproc{GrowOVER-Dialogue}. 
Since chat history is included in the prompts, it is harder for the classifier to gauge certainty based on documents and queries, so performance tends to suffer slightly.
Also, to show the actual effectiveness of the classifier, 
we separately calculate the average F1 / BLEU scores for adopted answers and not-adopted answers in the Decision Gate. The gap between the average F1 / BLEU score of adopted and not-adopted answers is approximately 25.0 for the QA task and 2.7 for the dialogue task. 
This significant gap indicates the classifier can predict the certainty and reliability of LLM. 

\vspace{-0.1cm}
\subsubsection{Results of Adaptive Re-Retrieval}
\vspace{-0.1cm}

We compare the accuracy of ARR and 
two baselines: i) search with only query ($Q$) and ii) always append ${y_\mathrm{LLM}}$ to $Q$ ($Q$:$y_\mathrm{LLM}$), each of which are using $sim(\mathbf{E}(Q), \mathbf{E}(D))$ and $sim(\mathbf{E}([Q$:$y_\mathrm{LLM}]), \mathbf{E}(D))$, where $\omega = 0$ and $\omega = 1$ in equation~(\ref{eqn:score}), respectively.
The gap between ARR and the former implies the degree to which reliable $y_\mathrm{LLM}$ improves retrieval relevance, while the gap between ARR and the latter represents the degree to which ARR ignores potentially incorrect $y_\mathrm{LLM}$. We exclude the first retrieved documents from re-retrieval to avoid using duplicate retrievals. The results are shown in table~\ref{tab:arr_sentbert}. 
In the QA task, ARR improves $Q$ and $Q$:$y_\mathrm{LLM}$ by approximately 1.2 and 0.3, respectively, while in the dialogue task, the performance enhances by 1.0 and 0.2, respectively.
This result indicates that the output of LLM with \textit{reliable} probability can aid in re-retrieval. 
Moreover, the effect of retrieval relevance on the final answer is demonstrated in the subsequent end-to-end results.

\begin{table}[t!]
\begin{center}
\renewcommand{\arraystretch}{1.2}
\resizebox{\linewidth}{!}{
{\footnotesize
\begin{tabular}{p{2.4cm} >{\centering\arraybackslash}p{1.1cm} >{\centering\arraybackslash}p{1.1cm} >{\centering\arraybackslash}p{1.1cm} >{\centering\arraybackslash}p{1.1cm}}
\Xhline{2\arrayrulewidth}
 & 9 & 10 & 11 & 12\\ 
\hline
\textbf{\textproc{New}} \\
\hline
    Vanilla & 13.4 & 14.1 & 14.7 & 14.1 \\
    Self-RAG & 23.6 & 22.6 & 23.2 & 22.7 \\
    RaLM & 38.2 & 36.8 & 37.0 & 37.1 \\
    RaLM-CP$^\dagger$ & \underline{39.2} & \underline{38.3} & 37.6 & 37.7 \\
    RaLM-$D^{*}$ \textnormal{(Ours)} & 37.9 & 38.1 & \underline{38.3} & \underline{38.1} \\
    RiLM \text{(Ours)}& \textbf{39.4} & \textbf{39.4} & \textbf{39.7} & \textbf{39.2} \\
\hline
\textbf{\textproc{Changed}} \\ 
\hline
    Vanilla & 6.1 & 5.6 & 7.0 & 5.3 \\
    Self-RAG & 18.2 & 19.2 & 17.7 & 20.1 \\
    RaLM & 24.9 & 26.5 & 33.2 & 28.5 \\
    RaLM-CP$^\dagger$ & \underline{26.0} & \textbf{29.0} & 33.6 & 29.6 \\
    RaLM-$D^{*}$ \textnormal{(Ours)} & 25.1 & \underline{27.7} & \underline{37.5} & \underline{29.9} \\
    RiLM \text{(Ours)}& \textbf{28.2} & \underline{27.7} & \textbf{38.3} & \textbf{30.4} \\
\hline
\textbf{\textproc{Unchanged}} \\ 
\hline
    Vanilla & 18.0 & 17.6 & 17.2 & 16.7 \\
    Self-RAG & 26.7 & 25.5 & 25.4 & 26.1 \\
    RaLM & 43.1 & 41.2 & 41.8 & 41.1 \\
    RaLM-CP$^\dagger$ & 44.0 & 43.3 & 42.7 & 42.0 \\
    RaLM-$D^{*}$ \text{(Ours)} & \underline{44.3} & \underline{43.7} & \underline{43.5} & \underline{42.9} \\
    RiLM \text{(Ours)}& \textbf{45.7} & \textbf{45.1} & \textbf{44.6} & \textbf{44.1} \\
\hline
\textbf{\textproc{All}} \\ 
\hline
    Vanilla & 17.4 & 17.1 & 16.8 & 16.4 \\
    Self-RAG & 26.3 & 25.1 & 25.1 & 25.7 \\
    RaLM & 42.5 & 40.5 & 41.2 & 40.6 \\
    RaLM-CP$^\dagger$ & 43.4 & 42.6 & 42.1 & 41.5 \\
    RaLM-$D^{*}$ \text{(Ours)} & \underline{43.5} & \underline{42.9} & \underline{42.8} & \underline{42.3} \\
    RiLM \text{(Ours)}& \textbf{44.9} & \textbf{44.2} & \textbf{44.0} & \textbf{43.5} \\
\Xhline{2\arrayrulewidth}
\multicolumn{5}{l}{\scriptsize $^\dagger$continuously pre-trained language model}
\end{tabular}
}
}
\vspace{-0.3cm}
\caption{\small F1 scores of \textproc{GrowOVER-QA} on each month. }
\label{tab:qa_result_sentbert}
\vspace{-0.65cm}
\end{center}
\end{table}

\vspace{-0.1cm}
\subsubsection{Results of \textproc{GrowOVER-QA}}

Table~\ref{tab:qa_result_sentbert} shows the performance on \textproc{GrowOVER-QA} from September to December. It displays the F1 score for \textproc{New}, \textproc{Changed}, and \textproc{Unchanged} QAs, as well as \textproc{All} types.
When comparing Vanilla to other baselines, it's clear that the retrieval significantly enhances performance, highlighting its crucial role in open-domain tasks.
Self-RAG shows modest improvement over Vanilla but underperforms compared to other retrieval-augmented models.
Also, in general, RiLM demonstrates outstanding performance over other baselines.
For \textproc{New}, RiLM improves RaLM from 1.2 in September to 2.7 in November. It shows a higher F1 score over all months than RaLM-CP, even though RiLM is not continuously trained.
For \textproc{Changed}, our method outperforms other baselines, except for October, when RaLM-CP exceeds our method by 1.3.
Nevertheless, our RiLM shows much more robust performance across the other months, with improvements ranging from 0.8 to 4.7. 
Moreover, for \textproc{Unchanged} and \textproc{All}, our method surpasses the performance of other baselines. 
However, over several months, all baselines show performance degradation over months, indicating the need for the model update in the future.

\vspace{-0.1cm}
\subsubsection{Results of \textproc{GrowOVER-Dialogue}}
\vspace{-0.05cm}

\begin{table}[t!]
\begin{center}
\renewcommand{\arraystretch}{1.2}
\resizebox{\linewidth}{!}{
{\footnotesize
\begin{tabular}{p{2.4cm} >{\centering\arraybackslash}p{1.1cm} >{\centering\arraybackslash}p{1.1cm} >{\centering\arraybackslash}p{1.1cm} >{\centering\arraybackslash}p{1.1cm}}
\Xhline{2\arrayrulewidth}
 & 9 & 10 & 11 & 12\\ 
\hline
\textbf{\textproc{New}} \\ 
\hline
    Vanilla & 0.85 & 0.81 & 0.84 & 0.88 \\
    Self-RAG & 2.37 & 2.29 & 2.36 & 2.21 \\
    RaLM & 4.98 & 5.08 & 5.06 & 4.76 \\
    RaLM-CP$^\dagger$ & 5.06 & 5.04 & 5.08 & 4.86 \\
    RaLM-$D^{*}$ \text{(Ours)} & \underline{5.27} & \underline{5.21} & \underline{5.42} & \underline{5.07} \\
    RiLM \text{(Ours)} & \textbf{5.36} & \textbf{5.27} & \textbf{5.51} & \textbf{5.15} \\
\hline
\textbf{\textproc{Changed}} \\ 
\hline
    Vanilla & 1.58 & 2.68 & 1.40 & 1.87 \\ 
    Self-RAG & 4.31 & 3.74 & 3.28 & 5.00 \\
    RaLM & 5.09 & 6.25 & \textbf{6.89} & 6.30 \\
    RaLM-CP$^\dagger$ & 6.11 & 6.98 & \underline{6.49} & 6.36 \\
    RaLM-$D^{*}$ \text{(Ours)} & \underline{6.60} & \underline{7.19} & 6.01 & \underline{6.38} \\
    RiLM \text{(Ours)} & \textbf{7.26} & \textbf{7.67} & 6.05 & \textbf{6.64} \\
\hline
\textbf{\textproc{Unchanged}} \\ 
\hline
    Vanilla & 1.13 & 1.12 & 1.10 & 1.11 \\
    Self-RAG & 2.58 & 2.56 & 2.32 & 2.49 \\
    RaLM & 4.42 & 4.41 & 4.44 & 4.45 \\
    RaLM-CP$^\dagger$ & 4.40 & 4.43 & 4.43 & 4.45 \\
    RaLM-$D^{*}$ \text{(Ours)} & \underline{4.65} & \underline{4.67} & \underline{4.69} & \underline{4.69} \\
    RiLM \text{(Ours)} & \textbf{4.68} & \textbf{4.69} & \textbf{4.71} & \textbf{4.71} \\
\hline
\textbf{\textproc{All}} \\ 
\hline
    Vanilla & 1.12 & 1.11 & 1.09 & 1.10 \\
    Self-RAG & 2.58 & 2.55 & 2.32 & 2.48 \\
    RaLM & 4.44 & 4.44 & 4.46 & 4.46 \\
    RaLM-CP$^\dagger$ & 4.43 & 4.45 & 4.46 & 4.47 \\
    RaLM-$D^{*}$ \text{(Ours)} & \underline{4.68} & \underline{4.69} & \underline{4.72} & \underline{4.70} \\
    RiLM \text{(Ours)} & \textbf{4.70} & \textbf{4.72} & \textbf{4.74} & \textbf{4.73} \\
\Xhline{2\arrayrulewidth}
\multicolumn{5}{l}{\scriptsize $^\dagger$continuously pre-trained language model}
\end{tabular}
}
}
\vspace{-0.3cm}
\caption{\footnotesize BLEU scores of \textproc{GrowOVER-Dialogue} on each month.} 
\label{tab:dialogue}
\vspace{-0.8cm}
\end{center}
\end{table}

Table~\ref{tab:dialogue} displays the consequences on \textproc{GrowOVER-Dialogue} from September to December.
Similar to QA task, RiLM generally exhibits superior performance over other baselines.
For instance, RiLM improves upon RaLM-CP by an average of 0.3 for \textproc{New} and 0.4 for \textproc{Changed}. Also, RiLM enhances performance on both the \textproc{Unchanged} and \textproc{All} datasets, indicating its overall effectiveness.
However, for \textproc{Changed} in November, not only RiLM but also RaLM-$D^{*}$ underperforms compared to RaLM. 
This performance drop only occurs on \textproc{Changed}, which may indicate the limitation in predicting \textit{misleading} cases. From the consistent performance improvements on \textproc{New} and \textproc{Unchanged}, we can infer that LLM can identify what it knows and what it doesn't know. On the other hand, it occasionally has difficulty assuming that its answer might be wrong and verifying it. 
\vspace{-0.2cm}

\subsubsection{Label-Based Analysis}
\vspace{-0.1cm}

In addition to the monthly results, we also report the average experimental results for each label across the months. Table~\ref{tab:arr_sentbert_label} shows ARR accuracy for \textproc{New}, \textproc{Changed}, and \textproc{Unchanged} labels.
For the \textproc{Changed} label in the QA task, $Q$:$y_\mathrm{LLM}$ performs 1.6 points lower than $Q$, indicating that relying entirely on misleading $y_\mathrm{LLM}$ can significantly degrade performance. Conversely, in the dialogue task, misleading $y_\mathrm{LLM}$ for \textproc{Changed} improves performance by 1.2 points. This suggests that the detailed sentences generated in the dialogue task benefit from common knowledge between the old and new data, aiding re-retrieval. For \textproc{New}, the LLM struggles to generate detailed answers, resulting in a small gain of 0.1 points. For \textproc{Unchanged} in both tasks, $y_\mathrm{LLM}$ improves performance by 1.4 and 0.8 points, respectively, though still lower than ARR. Overall, ARR demonstrates robust and improved performance in all scenarios.

To analyze errors corrected by Decision Gate and ARR, we report the performance at each stage of the RiLM pipeline, averaged by label type, as shown in Table~\ref{tab:clf_sentbert_label}. This includes results for answers adopted by the decision gate, not adopted answers before and after ARR and after the final selection, and the final aggregated result. The classifier performs well across all types, as evidenced by the gap between adopted and not-adopted answers before ARR. The efficiency of ARR and the decision gate is highlighted by the performance improvements from before ARR to after the final selection. Although performance decreases after ARR due to the exclusion of top-k documents from re-retrieval, the final selection results for not-adopted answers improve, indicating the classifier's effectiveness in selecting more reliable answers. In the QA task, F1 scores for not-adopted answers improve by over 2.0 points across all types. In the dialogue task, BLEU scores significantly improve for \textproc{New} and \textproc{Changed} by 1.0 and 0.6 points, respectively, with little improvement in \textproc{Unchanged}.

\begin{table}[t!]
\begin{center}
\renewcommand{\arraystretch}{1.2}
\resizebox{\linewidth}{!}{
{\footnotesize
\begin{tabular}{p{4.2cm} >{\centering\arraybackslash}p{0.8cm} >{\centering\arraybackslash}p{1.1cm} >{\centering\arraybackslash}p{1.4cm}}
\Xhline{2\arrayrulewidth}
 & \textproc{New} & \textproc{Changed} & \textproc{Unchanged} \\  
\hline
\textbf{\textproc{GrowOVER-QA}} \\ 
\hline
    $Q$ & 11.1 & 12.2 & 13.2 \\
    $Q$:$y_\mathrm{LLM}$ & \textbf{12.2} & 10.6 & 14.6 \\
    ARR & \textbf{12.2} & \textbf{12.6} & \textbf{14.7}  \\
\hline
\textbf{\textproc{GrowOVER-Dialogue}} \\ 
\hline
    $Q$ & 10.5 & 11.8 & 10.6 \\
    $Q$:$y_\mathrm{LLM}$ & 10.6 & 13.0 & 11.4 \\
    ARR & \textbf{11.0} & \textbf{13.7} & \textbf{11.6} \\
\Xhline{2\arrayrulewidth}
\end{tabular}
}
}
\vspace{-0.2cm}
\caption{\footnotesize Accuracy of ARR for each label.}
\label{tab:arr_sentbert_label}
\vspace{-0.4cm}
\end{center}
\end{table}

\begin{table}[t!]
\begin{center}
\renewcommand{\arraystretch}{1.2}
\resizebox{\linewidth}{!}{
{\footnotesize
\begin{tabular}{p{4cm} >{\centering\arraybackslash}p{1.1cm} >{\centering\arraybackslash}p{1.2cm} >{\centering\arraybackslash}p{1.4cm}}
\Xhline{2\arrayrulewidth}
 & \textproc{New} & \textproc{Changed} & \textproc{Unchanged}\\ 
\hline
\textbf{\textproc{GrowOVER-QA}} \\ 
\hline
    Adopted & 50.3 & 39.3 & 52.4 \\
    Not Adopted - Before ARR & 26.0 & 20.0 & 27.3 \\
    Not Adopted - After ARR & 19.6 & 11.9 & 21.9 \\
    Not Adopted - Final Selection & 28.7 & 22.0 & 31.0 \\
    Average & 39.4 & 31.2 & 44.9 \\
\hline
\textbf{\textproc{GrowOVER-Dialogue}} \\ 
\hline
    Adopted & 7.02 & 8.37 & 6.08\\
    Not Adopted - Before ARR & 3.97 & 5.33 & 3.41 \\
    Not Adopted - After ARR & 1.19 & 2.62 & 1.30 \\
    Not Adopted - Final Selection & 4.09 & 5.96 & 3.45 \\
    Average & 5.32 & 6.90 & 4.70 \\
\Xhline{2\arrayrulewidth}
\end{tabular}
}
}
\vspace{-0.2cm}
\caption{\footnotesize F1 and BLEU scores for each label in each process of the pipeline.}
\label{tab:clf_sentbert_label}
\vspace{-0.6cm}
\end{center}
\end{table}

\vspace{-0.cm}
\section{Conclusion}
\vspace{-0.2cm}

To evaluate whether LLMs can adapt to the fast-evolving world knowledge, we propose \textproc{GrowOVER-QA} and \textproc{GrowOVER-Dialogue}. Our benchmarks annotate the evidence text and introduce dialogue tasks to evaluate retriever-augmented RaLM comprehensively. Furthermore, we suggest RiLM, an interactive retriever-generator framework by simply training a classifier for LLM to predict reliability itself. 
Through rigorous experiments, we show that our method can be on par with or surpass continuously pre-trained LLMs even without pre-training. However, even with retrieval, we observe that the performance degrades over time. Thus, we hope our benchmarks can be valuable resources to detect when to update the retriever or LLM in future work. Additionally, we anticipate further research into optimizing the use of retrievers to reduce the frequency of model updates.

\section*{Limitations}

We highlight a few considerations for readers regarding potential limitations. 
We rely on various models to label each sentence in Wikipedia snapshots. Although we thoroughly designed the sentence labeling process to label sentences accurately, it can occasionally be faulty. To address this limitation, we append the MTurk study in the Appendix, showing the results are within acceptable bounds. 
Also, our dataset is primarily built from knowledge based on single articles, which may restrict its effectiveness for tasks that necessitate combining information from multiple sources.
Further research would be beneficial to generate benchmarks that enable the evaluation of frameworks handling complex reasoning tasks with a time-sensitive nature.
Moreover, the information extracted from Wikipedia may not promptly reflect real-world knowledge updates. This means that knowledge modifications in the real world might not be immediately mirrored in the dataset, resulting in some degree of outdated or inaccurate information.
Finally, the characteristics of our dataset may be influenced by the features of GPT-4 that we utilized during its generation, as well as the prompts we used. To enhance the effectiveness and mitigate bias in datasets, further research should focus on generating datasets for a variety of using Large Language Models employed across diverse fields, aligned with a wide range of prompts.

\section*{Ethics Statement}
We have manually reevaluated the dataset we created to ensure it is free of any potential for discrimination, human rights violations, bias, exploitation, and any other ethical concerns.

\section*{Acknowledgments}
We sincerely thank Jaewoo Ahn, Soochan Lee, Yeda Song, Heeseung Yun, Junik Bae, and other anonymous reviewers for their valuable comments. This work was supported by 
Institute of Information \& communications Technology Planning \& Evaluation (IITP) grant funded by the Korea government (MSIT) (No.~RS-2022-II220156, Fundamental research on continual meta-learning for quality enhancement of casual videos and their 3D metaverse transformation), 
the SNU-Global Excellence Research Center establishment project, 
the National Research Foundation of Korea (NRF) grant funded by the Korea government (MSIT) (No.~2023R1A2C2005573),
Basic Science Research Program through the National Research Foundation of Korea(NRF) funded by the Ministry of Education(RS-2023-00274280), 
and Institute of Information \& communications Technology Planning \& Evaluation (IITP) grant funded by the Korea government (MSIT) (No.~RS-2021-II211343, Artificial Intelligence Graduate School Program (Seoul National University)).

\bibliography{custom}

\begin{thebibliography}{42}
\providecommand{\natexlab}[1]{#1}

\bibitem[{Asai et~al.(2023)Asai, Wu, Wang, Sil, and Hajishirzi}]{asai2023self}
Akari Asai, Zeqiu Wu, Yizhong Wang, Avirup Sil, and Hannaneh Hajishirzi. 2023.
\newblock Self-rag: Learning to retrieve, generate, and critique through self-reflection.
\newblock In \emph{The Twelfth International Conference on Learning Representations}.

\bibitem[{Berant et~al.(2013)Berant, Chou, Frostig, and Liang}]{berant2013semantic}
Jonathan Berant, Andrew Chou, Roy Frostig, and Percy Liang. 2013.
\newblock Semantic parsing on freebase from question-answer pairs.
\newblock In \emph{Proceedings of the 2013 {C}onference on {E}mpirical {M}ethods in {N}atural {L}anguage {P}rocessing}, pages 1533--1544.

\bibitem[{Chan et~al.(2021)Chan, Liu, Li, Zhang, Zhao, Shi, and Yan}]{chan2021enhancing}
Zhangming Chan, Lemao Liu, Juntao Li, Haisong Zhang, Dongyan Zhao, Shuming Shi, and Rui Yan. 2021.
\newblock Enhancing the open-domain dialogue evaluation in latent space.
\newblock In \emph{Findings of the Association for Computational Linguistics: ACL-IJCNLP 2021}, pages 4889--4900.

\bibitem[{Chen et~al.(2020)Chen, Hou, Cui, Che, Liu, and Yu}]{chen2020recall}
Sanyuan Chen, Yutai Hou, Yiming Cui, Wanxiang Che, Ting Liu, and Xiangzhan Yu. 2020.
\newblock \href {https://arxiv.org/abs/2004.12651} {{Recall and Learn: Fine-tuning Deep Pretrained Language Models with Less Forgetting}}.
\newblock \emph{Preprint}, arXiv:2004.12651.

\bibitem[{Chen and Liu(2018)}]{chen2018lifelong}
Zhiyuan Chen and Bing Liu. 2018.
\newblock \emph{Lifelong machine learning}, volume~1.
\newblock Springer.

\bibitem[{Dhingra et~al.(2022)Dhingra, Cole, Eisenschlos, Gillick, Eisenstein, and Cohen}]{dhingra2022time}
Bhuwan Dhingra, Jeremy~R Cole, Julian~Martin Eisenschlos, Daniel Gillick, Jacob Eisenstein, and William~W Cohen. 2022.
\newblock Time-aware language models as temporal knowledge bases.
\newblock \emph{Transactions of the Association for Computational Linguistics}, 10:257--273.

\bibitem[{Dinan et~al.(2018)Dinan, Roller, Shuster, Fan, Auli, and Weston}]{dinan2018wizard}
Emily Dinan, Stephen Roller, Kurt Shuster, Angela Fan, Michael Auli, and Jason Weston. 2018.
\newblock Wizard of wikipedia: Knowledge-powered conversational agents.
\newblock In \emph{International Conference on Learning Representations}.

\bibitem[{Guu et~al.(2020)Guu, Lee, Tung, Pasupat, and Chang}]{guu2020retrieval}
Kelvin Guu, Kenton Lee, Zora Tung, Panupong Pasupat, and Mingwei Chang. 2020.
\newblock Retrieval augmented language model pre-training.
\newblock In \emph{International {C}onference on {M}achine {L}earning}, pages 3929--3938. PMLR.

\bibitem[{He et~al.(2021)He, Liu, Cho, Ott, Liu, Glass, and Peng}]{he2021analyzing}
Tianxing He, Jun Liu, Kyunghyun Cho, Myle Ott, Bing Liu, James Glass, and Fuchun Peng. 2021.
\newblock \href {https://arxiv.org/abs/1910.07117} {Analyzing the forgetting problem in the pretrain-finetuning of dialogue response models}.
\newblock \emph{Preprint}, arXiv:1910.07117.

\bibitem[{Hoffart et~al.(2011)Hoffart, Yosef, Bordino, F{\"u}rstenau, Pinkal, Spaniol, Taneva, Thater, and Weikum}]{hoffart2011robust}
Johannes Hoffart, Mohamed~Amir Yosef, Ilaria Bordino, Hagen F{\"u}rstenau, Manfred Pinkal, Marc Spaniol, Bilyana Taneva, Stefan Thater, and Gerhard Weikum. 2011.
\newblock Robust disambiguation of named entities in text.
\newblock In \emph{Proceedings of the 2011 {C}onference on {E}mpirical {M}ethods in {N}atural {L}anguage {P}rocessing}, pages 782--792.

\bibitem[{Hu et~al.(2021)Hu, Shen, Wallis, Allen-Zhu, Li, Wang, Wang, and Chen}]{hu2021lora}
Edward~J. Hu, Yelong Shen, Phillip Wallis, Zeyuan Allen-Zhu, Yuanzhi Li, Shean Wang, Lu~Wang, and Weizhu Chen. 2021.
\newblock \href {https://arxiv.org/abs/2106.09685} {Lora: Low-rank adaptation of large language models}.
\newblock \emph{Preprint}, arXiv:2106.09685.

\bibitem[{Izacard et~al.(2021)Izacard, Caron, Hosseini, Riedel, Bojanowski, Joulin, and Grave}]{izacard2021unsupervised}
Gautier Izacard, Mathilde Caron, Lucas Hosseini, Sebastian Riedel, Piotr Bojanowski, Armand Joulin, and Edouard Grave. 2021.
\newblock Unsupervised dense information retrieval with contrastive learning.
\newblock \emph{arXiv preprint arXiv:2112.09118}.

\bibitem[{Izacard et~al.(2022)Izacard, Lewis, Lomeli, Hosseini, Petroni, Schick, Dwivedi-Yu, Joulin, Riedel, and Grave}]{izacard2022few}
Gautier Izacard, Patrick Lewis, Maria Lomeli, Lucas Hosseini, Fabio Petroni, Timo Schick, Jane Dwivedi-Yu, Armand Joulin, Sebastian Riedel, and Edouard Grave. 2022.
\newblock Few-shot learning with retrieval augmented language models.
\newblock \emph{arXiv preprint arXiv:2208.03299}.

\bibitem[{Jang et~al.(2022)Jang, Ye, Lee, Yang, Shin, Han, Kim, and Seo}]{jang2022temporalwiki}
Joel Jang, Seonghyeon Ye, Changho Lee, Sohee Yang, Joongbo Shin, Janghoon Han, Gyeonghun Kim, and Minjoon Seo. 2022.
\newblock Temporalwiki: A lifelong benchmark for training and evaluating ever-evolving language models.
\newblock \emph{arXiv preprint arXiv:2204.14211}.

\bibitem[{Jang et~al.(2021)Jang, Ye, Yang, Shin, Han, Gyeonghun, Choi, and Seo}]{jang2021towards}
Joel Jang, Seonghyeon Ye, Sohee Yang, Joongbo Shin, Janghoon Han, KIM Gyeonghun, Stanley~Jungkyu Choi, and Minjoon Seo. 2021.
\newblock Towards continual knowledge learning of language models.
\newblock In \emph{International Conference on Learning Representations}.

\bibitem[{Jiang et~al.(2023)Jiang, Xu, Gao, Sun, Liu, Dwivedi-Yu, Yang, Callan, and Neubig}]{jiang2023active}
Zhengbao Jiang, Frank~F Xu, Luyu Gao, Zhiqing Sun, Qian Liu, Jane Dwivedi-Yu, Yiming Yang, Jamie Callan, and Graham Neubig. 2023.
\newblock Active retrieval augmented generation.
\newblock In \emph{Proceedings of the 2023 Conference on Empirical Methods in Natural Language Processing}, pages 7969--7992.

\bibitem[{Johnson et~al.(2019)Johnson, Douze, and J{\'e}gou}]{johnson2019billion}
Jeff Johnson, Matthijs Douze, and Herv{\'e} J{\'e}gou. 2019.
\newblock Billion-scale similarity search with {GPUs}.
\newblock \emph{IEEE Transactions on Big Data}, 7(3):535--547.

\bibitem[{Joshi et~al.(2017)Joshi, Choi, Weld, and Zettlemoyer}]{joshi2017triviaqa}
Mandar Joshi, Eunsol Choi, Daniel~S Weld, and Luke Zettlemoyer. 2017.
\newblock Triviaqa: A large scale distantly supervised challenge dataset for reading comprehension.
\newblock In \emph{Proceedings of the 55th Annual Meeting of the Association for Computational Linguistics (Volume 1: Long Papers)}, pages 1601--1611.

\bibitem[{Kasai et~al.(2022)Kasai, Sakaguchi, Takahashi, Bras, Asai, Yu, Radev, Smith, Choi, and Inui}]{kasai2022realtime}
Jungo Kasai, Keisuke Sakaguchi, Yoichi Takahashi, Ronan~Le Bras, Akari Asai, Xinyan Yu, Dragomir Radev, Noah~A Smith, Yejin Choi, and Kentaro Inui. 2022.
\newblock Realtime qa: What's the answer right now?
\newblock \emph{arXiv preprint arXiv:2207.13332}.

\bibitem[{Kim et~al.(2023)Kim, Yoon, Ye, Hwang, and Yun}]{kim2023carpe}
Yujin Kim, Jaehong Yoon, Seonghyeon Ye, Sung~Ju Hwang, and Se-young Yun. 2023.
\newblock Carpe diem: On the evaluation of world knowledge in lifelong language models.
\newblock \emph{arXiv preprint arXiv:2311.08106}.

\bibitem[{Kwiatkowski et~al.(2019)Kwiatkowski, Palomaki, Redfield, Collins, Parikh, Alberti, Epstein, Polosukhin, Devlin, Lee et~al.}]{kwiatkowski2019natural}
Tom Kwiatkowski, Jennimaria Palomaki, Olivia Redfield, Michael Collins, Ankur Parikh, Chris Alberti, Danielle Epstein, Illia Polosukhin, Jacob Devlin, Kenton Lee, et~al. 2019.
\newblock Natural questions: a benchmark for question answering research.
\newblock \emph{Transactions of the Association for Computational Linguistics}, 7:453--466.

\bibitem[{Lazaridou et~al.(2021)Lazaridou, Kuncoro, Gribovskaya, Agrawal, Liska, Terzi, Gimenez, de~Masson~d'Autume, Kocisky, Ruder et~al.}]{lazaridou2021mind}
Angeliki Lazaridou, Adhi Kuncoro, Elena Gribovskaya, Devang Agrawal, Adam Liska, Tayfun Terzi, Mai Gimenez, Cyprien de~Masson~d'Autume, Tomas Kocisky, Sebastian Ruder, et~al. 2021.
\newblock Mind the gap: Assessing temporal generalization in neural language models.
\newblock \emph{Advances in Neural Information Processing Systems}, 34:29348--29363.

\bibitem[{Levy et~al.(2017)Levy, Seo, Choi, and Zettlemoyer}]{levy2017zero}
Omer Levy, Minjoon Seo, Eunsol Choi, and Luke Zettlemoyer. 2017.
\newblock Zero-shot relation extraction via reading comprehension.
\newblock In \emph{Proceedings of the 21st Conference on Computational Natural Language Learning (CoNLL 2017)}, pages 333--342.

\bibitem[{Lewis et~al.(2020)Lewis, Perez, Piktus, Petroni, Karpukhin, Goyal, K{\"u}ttler, Lewis, Yih, Rockt{\"a}schel et~al.}]{lewis2020retrieval}
Patrick Lewis, Ethan Perez, Aleksandra Piktus, Fabio Petroni, Vladimir Karpukhin, Naman Goyal, Heinrich K{\"u}ttler, Mike Lewis, Wen-tau Yih, Tim Rockt{\"a}schel, et~al. 2020.
\newblock Retrieval-augmented generation for knowledge-intensive nlp tasks.
\newblock \emph{Advances in Neural Information Processing Systems}, 33:9459--9474.

\bibitem[{Liska et~al.(2022)Liska, Kocisky, Gribovskaya, Terzi, Sezener, Agrawal, Cyprien De~Masson, Scholtes, Zaheer, Young et~al.}]{liska2022streamingqa}
Adam Liska, Tomas Kocisky, Elena Gribovskaya, Tayfun Terzi, Eren Sezener, Devang Agrawal, D’Autume Cyprien De~Masson, Tim Scholtes, Manzil Zaheer, Susannah Young, et~al. 2022.
\newblock Streamingqa: A benchmark for adaptation to new knowledge over time in question answering models.
\newblock In \emph{International Conference on Machine Learning}, pages 13604--13622. PMLR.

\bibitem[{Liu et~al.(2019)Liu, Ott, Goyal, Du, Joshi, Chen, Levy, Lewis, Zettlemoyer, and Stoyanov}]{liu2019roberta}
Yinhan Liu, Myle Ott, Naman Goyal, Jingfei Du, Mandar Joshi, Danqi Chen, Omer Levy, Mike Lewis, Luke Zettlemoyer, and Veselin Stoyanov. 2019.
\newblock Roberta: A robustly optimized bert pretraining approach.
\newblock \emph{arXiv preprint arXiv:1907.11692}.

\bibitem[{Longpre et~al.(2021)Longpre, Perisetla, Chen, Ramesh, DuBois, and Singh}]{longpre2021entity}
Shayne Longpre, Kartik Perisetla, Anthony Chen, Nikhil Ramesh, Chris DuBois, and Sameer Singh. 2021.
\newblock Entity-based knowledge conflicts in question answering.
\newblock \emph{arXiv preprint arXiv:2109.05052}.

\bibitem[{Luu et~al.(2021)Luu, Khashabi, Gururangan, Mandyam, and Smith}]{luu2021time}
Kelvin Luu, Daniel Khashabi, Suchin Gururangan, Karishma Mandyam, and Noah~A Smith. 2021.
\newblock Time waits for no one! analysis and challenges of temporal misalignment.
\newblock \emph{arXiv preprint arXiv:2111.07408}.

\bibitem[{Margatina et~al.(2023)Margatina, Wang, Vyas, John, Benajiba, and Ballesteros}]{margatina2023dynamic}
Katerina Margatina, Shuai Wang, Yogarshi Vyas, Neha~Anna John, Yassine Benajiba, and Miguel Ballesteros. 2023.
\newblock Dynamic benchmarking of masked language models on temporal concept drift with multiple views.
\newblock In \emph{Proceedings of the 17th Conference of the European Chapter of the Association for Computational Linguistics}, pages 2873--2890.

\bibitem[{Papineni et~al.(2002)Papineni, Roukos, Ward, and Zhu}]{papineni2002bleu}
Kishore Papineni, Salim Roukos, Todd Ward, and Wei-Jing Zhu. 2002.
\newblock Bleu: a method for automatic evaluation of machine translation.
\newblock In \emph{Proceedings of the 40th annual meeting of the Association for Computational Linguistics}, pages 311--318.

\bibitem[{Petroni et~al.(2021)Petroni, Piktus, Fan, Lewis, Yazdani, De~Cao, Thorne, Jernite, Karpukhin, Maillard et~al.}]{petroni2021kilt}
Fabio Petroni, Aleksandra Piktus, Angela Fan, Patrick Lewis, Majid Yazdani, Nicola De~Cao, James Thorne, Yacine Jernite, Vladimir Karpukhin, Jean Maillard, et~al. 2021.
\newblock Kilt: a benchmark for knowledge intensive language tasks.
\newblock In \emph{Proceedings of the 2021 Conference of the North American Chapter of the Association for Computational Linguistics: Human Language Technologies}, pages 2523--2544.

\bibitem[{Ram et~al.(2023)Ram, Levine, Dalmedigos, Muhlgay, Shashua, Leyton-Brown, and Shoham}]{ram2023context}
Ori Ram, Yoav Levine, Itay Dalmedigos, Dor Muhlgay, Amnon Shashua, Kevin Leyton-Brown, and Yoav Shoham. 2023.
\newblock In-context retrieval-augmented language models.
\newblock \emph{arXiv preprint arXiv:2302.00083}.

\bibitem[{Reimers and Gurevych(2019)}]{reimers2019sentence}
Nils Reimers and Iryna Gurevych. 2019.
\newblock Sentence-bert: Sentence embeddings using siamese bert-networks.
\newblock \emph{arXiv preprint arXiv:1908.10084}.

\bibitem[{Saxena et~al.(2021)Saxena, Chakrabarti, and Talukdar}]{saxena2021question}
Apoorv Saxena, Soumen Chakrabarti, and Partha Talukdar. 2021.
\newblock Question answering over temporal knowledge graphs.
\newblock \emph{arXiv preprint arXiv:2106.01515}.

\bibitem[{Shi et~al.(2023)Shi, Min, Yasunaga, Seo, James, Lewis, Zettlemoyer, and Yih}]{shi2023replug}
Weijia Shi, Sewon Min, Michihiro Yasunaga, Minjoon Seo, Rich James, Mike Lewis, Luke Zettlemoyer, and Wen-tau Yih. 2023.
\newblock Replug: Retrieval-augmented black-box language models.
\newblock \emph{arXiv preprint arXiv:2301.12652}.

\bibitem[{Thorne et~al.(2018)Thorne, Vlachos, Christodoulopoulos, and Mittal}]{thorne2018fever}
James Thorne, Andreas Vlachos, Christos Christodoulopoulos, and Arpit Mittal. 2018.
\newblock Fever: a large-scale dataset for fact extraction and verification.
\newblock In \emph{Proceedings of the 2018 Conference of the North American Chapter of the Association for Computational Linguistics: Human Language Technologies, Volume 1 (Long Papers)}, pages 809--819.

\bibitem[{Touvron et~al.(2023)Touvron, Martin, Stone, Albert, Almahairi, Babaei, Bashlykov, Batra, Bhargava, Bhosale et~al.}]{touvron2023llama}
Hugo Touvron, Louis Martin, Kevin Stone, Peter Albert, Amjad Almahairi, Yasmine Babaei, Nikolay Bashlykov, Soumya Batra, Prajjwal Bhargava, Shruti Bhosale, et~al. 2023.
\newblock Llama 2: Open foundation and fine-tuned chat models.
\newblock \emph{arXiv preprint arXiv:2307.09288}.

\bibitem[{Wang et~al.(2020)Wang, Tang, Duan, Wei, Huang, ji, Cao, Jiang, and Zhou}]{wang2020kadapter}
Ruize Wang, Duyu Tang, Nan Duan, Zhongyu Wei, Xuanjing Huang, Jianshu ji, Guihong Cao, Daxin Jiang, and Ming Zhou. 2020.
\newblock \href {https://arxiv.org/abs/2002.01808} {K-adapter: Infusing knowledge into pre-trained models with adapters}.
\newblock \emph{Preprint}, arXiv:2002.01808.

\bibitem[{Williams et~al.(2018)Williams, Nangia, and Bowman}]{williams-etal-2018-broad}
Adina Williams, Nikita Nangia, and Samuel Bowman. 2018.
\newblock \href {https://doi.org/10.18653/v1/N18-1101} {A broad-coverage challenge corpus for sentence understanding through inference}.
\newblock In \emph{Proceedings of the 2018 Conference of the North {A}merican Chapter of the Association for Computational Linguistics: Human Language Technologies, Volume 1 (Long Papers)}, pages 1112--1122, New Orleans, Louisiana. Association for Computational Linguistics.

\bibitem[{Xu et~al.(2023)Xu, Xie, Qin, Tao, and Wang}]{xu2023parameterefficient}
Lingling Xu, Haoran Xie, Si-Zhao~Joe Qin, Xiaohui Tao, and Fu~Lee Wang. 2023.
\newblock \href {https://arxiv.org/abs/2312.12148} {Parameter-efficient fine-tuning methods for pretrained language models: A critical review and assessment}.
\newblock \emph{Preprint}, arXiv:2312.12148.

\bibitem[{Yang et~al.(2018)Yang, Qi, Zhang, Bengio, Cohen, Salakhutdinov, and Manning}]{yang2018hotpotqa}
Zhilin Yang, Peng Qi, Saizheng Zhang, Yoshua Bengio, William Cohen, Ruslan Salakhutdinov, and Christopher~D Manning. 2018.
\newblock Hotpotqa: A dataset for diverse, explainable multi-hop question answering.
\newblock In \emph{Proceedings of the 2018 Conference on Empirical Methods in Natural Language Processing}, pages 2369--2380.

\bibitem[{Zhang and Choi(2021)}]{zhang-choi-2021-situatedqa}
Michael Zhang and Eunsol Choi. 2021.
\newblock \href {https://doi.org/10.18653/v1/2021.emnlp-main.586} {{S}ituated{QA}: Incorporating extra-linguistic contexts into {QA}}.
\newblock In \emph{Proceedings of the 2021 Conference on Empirical Methods in Natural Language Processing}, pages 7371--7387, Online and Punta Cana, Dominican Republic. Association for Computational Linguistics.

\end{thebibliography}

\newpage
\appendix
\clearpage
\section{Experimental Details}\label{expDetails}

\textbf{Dataset}. 
For dataset generation, we use OpenAI \textit{gpt-4-1106-preview} model and set the temperature as 0 and max\_new\_token as 256.
Also, we use Huggingface \textit{princeton-nlp/sup-simcse-roberta-large} for NLI task in the sentence labeling process. Additionally, when checking contradiction with GPT-4, we use \textit{gpt-4-1106-preview} and set the temperature as 0 and max\_tokens as 256.

\textbf{Database}. We use the Langchain document loader (RecursiveCharacterTextSplitter) to split each article in the Wikipedia snapshot into several passages. We set chunk\_size as 1500 and chunk\_overlap as 10 characters.

\textbf{Continual Pretraining of LLM}. For initial training on August, we train Llama2 for four epochs with a learning rate of 1e-06, a learning rate decay of 0.8, a cosine learning rate scheduler, an AdamW optimizer, and a batch size of 64, using FSDP (fully sharded data parallel). After that, for the c-RaLM baseline, we continuously pre-train the model with 12K articles on each month. We set all hyperparameters the same with the initial training except the epoch. To prevent catastrophic forgetting, we trained the model for only one epoch, following \citet{jang2022temporalwiki}.

\textbf{Classifier Training}. For the classifier, we trained a linear layer with dimensions (4096, 3). In both QA and dialogue response generation tasks, we set the learning rate as 0.0001, weight decay as 1e-07, and 10 epochs. We use a cosine learning rate scheduler. Also, we train the model using cross-entropy loss.

\textbf{Adaptive Re-Retrieval}. In both QA and dialogue tasks with SentBERT, we set the hyper-parameter \(\lambda\) as 2.0. In the other cases, we set the \(\lambda\) as 1.0.

\textbf{Answer Generation}. We set the max\_new\_token of Llama2 as 10 for QA and 50 for dialogue, and load the model using bfloat16. 

\textbf{Self-RAG}. We used short-form generation version with always\_retrieve mode, since all data points of \textproc{GrowOver} contain evidence texts. We used same retrievals as RiLM. 
\section{Additional Experimental Results}

\mycomment{
\subsection{Self-RAG}\label{app:self_rag}
We conduct evaluations using Self-RAG on \textproc{GrowOver-QA}. For a fair comparison, we evaluate the short-form generation of Self-RAG on GrowOVER-QA using the same retrieval settings as ours. The results averaged across all months are shown in the table~\ref{tab:qa_self_rag}.
Even though we freeze LLMs and train the classifier only, our RiLM outperforms Self-RAG by a significant margin. 

\begin{table}[H]
\begin{center}
\renewcommand{\arraystretch}{1.2}
\resizebox{\linewidth}{!}{
{\footnotesize
\begin{tabular}{p{2.4cm} >{\centering\arraybackslash}p{1.1cm} >{\centering\arraybackslash}p{1.1cm} >{\centering\arraybackslash}p{1.1cm} >{\centering\arraybackslash}p{1.1cm}}
\Xhline{2\arrayrulewidth}
 & 9 & 10 & 11 & 12\\ 
\hline
\textbf{\textproc{New}} \\ 
\hline
    Self-RAG & 23.6 & 22.6 & 23.2 & 22.7 \\
    RiLM \text{(Ours)}& \textbf{39.4} & \textbf{39.4} & \textbf{39.7} & \textbf{39.2} \\
\hline
\textbf{\textproc{Changed}} \\ 
\hline
    Self-RAG & 18.2 & 19.2 & 17.7 & 20.1 \\
    RiLM \text{(Ours)}& \textbf{28.2} & \textbf{27.7} & \textbf{38.3} & \textbf{30.4} \\
\hline
\textbf{\textproc{Unchanged}} \\ 
\hline
    Self-RAG & 26.7 & 25.5 & 25.4 & 26.1 \\
    RiLM \text{(Ours)}& \textbf{45.7} & \textbf{45.1} & \textbf{44.6} & \textbf{44.1} \\
\Xhline{2\arrayrulewidth}
\multicolumn{5}{l}{\scriptsize $^\dagger$continuously pretrained language model}
\end{tabular}
}
}
\caption{F1 score of \textproc{GrowOVER-QA} on each month using Self-RAG.}
\label{tab:qa_self_rag}
\end{center}
\end{table}
}

\subsection{Contriever}\label{addexpResults}
We additionally perform experiments with Contriever \citep{izacard2021unsupervised}, following \citet{shi2023replug} and \citet{izacard2022few}. We report classifier performance, ARR results, as well as end-to-end QA results. 
\subsubsection{Classifier Performance}
Table~\ref{tab:clf_contriever} shows the experimental results of the certainty classifier. Despite using the same certainty classifier as SentBERT, the accuracy of all cases is above 65.0, demonstrating its effectiveness. Also, across all months, F1 / BLEU scores for adopted and not-adopted answers show a significant gap. 

\begin{table}[H]
\begin{center}
\renewcommand{\arraystretch}{1.2}
\resizebox{\linewidth}{!}{
{\footnotesize
\begin{tabular}{p{4.0cm} >{\centering\arraybackslash}p{0.85cm} >{\centering\arraybackslash}p{0.85cm} >{\centering\arraybackslash}p{0.85cm} >{\centering\arraybackslash}p{0.85cm}}
\Xhline{2\arrayrulewidth}
 & 9 & 10 & 11 & 12\\ 
\hline
\textbf{\textproc{GrowOVER-QA}} \\ 
\hline
    Accuracy & 77.2 & 74.5 & 74.4 & 72.9 \\
    Average F1 (Adopted) & 28.4 & 27.7 & 27.4 & 27.1 \\
    Average F1 (Not-adopted) & 17.0 & 16.7 & 16.5 & 16.2 \\
    Average F1 (ALL) & 21.1 & 20.6 & 20.2 & 19.8  \\
\hline
\textbf{\textproc{GrowOVER-Dialogue}} \\ 
\hline
    Accuracy & 66.6 & 66.4 & 65.9 & 66.0 \\
    Average BLEU (Adopted) & 4.44 & 4.44 & 4.46 & 4.46 \\
    Average BLEU (Not-adopted) & 2.53 & 2.56 & 2.55 & 2.56 \\
    Average BLEU (ALL) & 2.57 & 2.57 & 2.57 & 2.59 \\
\Xhline{2\arrayrulewidth}
\end{tabular}
}
}
\caption{Accuracy and F1 scores of the classifier in the Decision Gate on each month.}
\label{tab:clf_contriever}
\end{center}
\end{table}

\subsubsection{Adaptive Re-Retrieval results}
As shown in Table~\ref{tab:arr_contriever}, ARR significantly improves retrieval relevance. In the QA task, performance increased by more than 3 times, and in the dialogue task, it increased by more than 2 times. 
\begin{table}[H]
\begin{center}
\renewcommand{\arraystretch}{1.2}
\resizebox{\linewidth}{!}{
{\footnotesize
\begin{tabular}{p{3.6cm} >{\centering\arraybackslash}p{0.85cm} >{\centering\arraybackslash}p{0.85cm} >{\centering\arraybackslash}p{0.85cm} >{\centering\arraybackslash}p{0.85cm}}
\Xhline{2\arrayrulewidth}
 & 9 & 10 & 11 & 12\\  
\hline
\textbf{\textproc{GrowOVER-QA}} \\ 
\hline
    Next top-k & 2.6 & 2.6 & 2.5 & 2.6\\
    ARR & \textbf{8.5} & \textbf{8.2} & \textbf{8.0} & \textbf{8.0}\\
\hline
\textbf{\textproc{GrowOVER-Dialogue}} \\ 
\hline
    Next top-k & 11.6 & 11.6 & 11.6 & 11.6 \\
    ARR & \textbf{27.1} & \textbf{27.6} & \textbf{27.3} & \textbf{30.7}\\
\Xhline{2\arrayrulewidth}
\end{tabular}
}
}
\caption{Adaptive Re-Retrieval relevance of Contriever compared to choosing next top-k documents on each month.}
\label{tab:arr_contriever}
\end{center}
\end{table}

\subsubsection{QA Results}
The results of \textproc{GrowOVER-QA} are presented in Table~\ref{tab:qa_contriever}. It demonstrates the robustness of our method. For \textproc{New}, \textproc{Unchanged}, and all types of QA, our RiLM shows the highest score across all months. Also, for \textproc{Changed}, RiLM surpasses the other baselines except in September. In September, the performance degrades after re-retrieval, which indicates the DG possibly struggles with selecting correct answers between two different answers. Nonetheless, across all the other months, RiLM improves RaLM-$D^{*}$ about by 3.0. Also, it significantly outperforms other baselines.

\begin{table}[H]
\begin{center}
\renewcommand{\arraystretch}{1.2}
\resizebox{\linewidth}{!}{
{\footnotesize
\begin{tabular}{p{2.4cm} >{\centering\arraybackslash}p{1.1cm} >{\centering\arraybackslash}p{1.1cm} >{\centering\arraybackslash}p{1.1cm} >{\centering\arraybackslash}p{1.1cm}}
\Xhline{2\arrayrulewidth}
 & 9 & 10 & 11 & 12\\ 
\hline
\textbf{\textproc{New}} \\ 
\hline
    Vanilla & 13.4 & 14.1 & 14.7 & 14.1 \\
    RaLM & 16.5 & 16.7 & 17.1 & 16.4 \\
    RaLM-CP$^\dagger$ & 17.0 & 17.0 & 17.3 & 16.8 \\
    RaLM-$D^{*}$ \text{(Ours)} & \underline{17.2} & \underline{17.9} & \underline{18.0} & \underline{17.0} \\
    RiLM \text{(Ours)} & \textbf{19.8} & \textbf{20.2} & \textbf{19.8} & \textbf{18.8} \\
\hline
\textbf{\textproc{Changed}} \\ 
\hline
    Vanilla & 6.1 & 5.6 & 7.0 & 5.3 \\
    RaLM & 8.1 & 10.7 & 9.0 & 9.2 \\
    RaLM-CP$^\dagger$ & 8.1 & \underline{11.1} & 9.1 & \underline{9.3} \\
    RaLM-$D^{*}$ \text{(Ours)} & \textbf{11.8} & 10.9 & \underline{10.5} & 8.8 \\
    RiLM \text{(Ours)} & \underline{11.4} & \textbf{13.8} & \textbf{13.5} & \textbf{12.6} \\
\hline
\textbf{\textproc{Unchanged}} \\ 
\hline
    Vanilla & 18.0 & 17.6 & 17.2 & 16.7 \\
    RaLM & 20.3 & 19.8 & 19.4 & 19.2 \\
    RaLM-CP$^\dagger$ & 20.9 & 20.3 & 19.9 & 19.6 \\
    RaLM-$D^{*}$ \text{(Ours)} & \underline{21.7} & \underline{21.0} & \underline{20.6} & \underline{20.3} \\
    RiLM \text{(Ours)} & \textbf{23.6} & \textbf{22.9} & \textbf{22.4} & \textbf{22.1} \\
\hline
\textbf{\textproc{All}} \\ 
\hline
    Vanilla & 17.4 & 17.1 & 16.8 & 16.4 \\
    RaLM & 19.8 & 19.3 & 19.0 & 18.8 \\
    RaLM-CP$^\dagger$ & 20.4 & 19.8 & 19.6 & 19.3 \\
    RaLM-$D^{*}$ \text{(Ours)} & \underline{21.1} & \underline{20.6} & \underline{20.2} & \underline{19.8} \\
    RiLM \text{(Ours)} & \textbf{23.1} & \textbf{22.5} & \textbf{22.1} & \textbf{21.7} \\
\Xhline{2\arrayrulewidth}
\multicolumn{5}{l}{\scriptsize $^\dagger$continuously pretrained language model}
\end{tabular}
}
}
\caption{F1 score of \textproc{GrowOVER-QA} on each month using Contriever.}
\label{tab:qa_contriever}
\end{center}
\end{table}
\newpage

\begin{table}[H]
\begin{center}
\renewcommand{\arraystretch}{1.2}
\resizebox{\linewidth}{!}{
{\footnotesize
\begin{tabular}{p{2.4cm} >{\centering\arraybackslash}p{1.1cm} >{\centering\arraybackslash}p{1.1cm} >{\centering\arraybackslash}p{1.1cm} >{\centering\arraybackslash}p{1.1cm}}
\Xhline{2\arrayrulewidth}
 & 9 & 10 & 11 & 12\\ 
\hline
\textbf{\textproc{New}} \\ 
\hline
    Vanilla & 0.85 & 0.81 & 0.84 & 0.88 \\
    RaLM & 3.28 & 3.11 & 3.22 & 2.95 \\
    RaLM-CP$^\dagger$ & \underline{3.30} & \underline{3.13} & \underline{3.25} & 2.94 \\
    RaLM-$D^{*}$ \text{(Ours)} & 3.26 & 3.07 & 3.20 & \textbf{3.10} \\
    RiLM \text{(Ours)} & \textbf{3.61} & \textbf{3.14} & \textbf{3.66} & \textbf{3.10} \\
\hline
\textbf{\textproc{Changed}} \\ 
\hline
    Vanilla & 1.58 & 2.68 & 1.40 & 1.87 \\
    RaLM & \textbf{5.40} & \textbf{5.39} & 3.79 & 4.63 \\
    RaLM-CP$^\dagger$ & \underline{4.74} & \underline{5.08} & \underline{4.06} & 4.64 \\
    RaLM-$D^{*}$ \text{(Ours)} & 3.94 & 4.66 & 3.92 & \textbf{5.29} \\
    RiLM \text{(Ours)} & 4.28 & 4.71 & \textbf{4.48} & \textbf{5.29} \\
\hline
\textbf{\textproc{Unchanged}} \\ 
\hline
    Vanilla & 1.13 & 1.12 & 1.10 & 1.11 \\
    RaLM & 2.50 & 2.53 & 2.52 & 2.54 \\
    RaLM-CP$^\dagger$ & 2.48 & 2.51 & 2.52 & 2.53 \\
    RaLM-$D^{*}$ \text{(Ours)} & \underline{2.54} & \underline{2.55} & \underline{2.54} & \underline{2.56} \\
    RiLM \text{(Ours)} & \textbf{2.66} & \textbf{2.56} & \textbf{2.64} & \textbf{2.58} \\
\hline
\textbf{\textproc{All}} \\ 
\hline
    Vanilla & 1.12 & 1.11 & 1.09 & 1.10 \\
    RaLM & 2.53 & 2.56 & 2.55 & 2.56 \\
    RaLM-CP$^\dagger$ & 2.51 & 2.54 & 2.55 & 2.55 \\
    RaLM-$D^{*}$ \text{(Ours)} & \underline{2.57} & \underline{2.57} & \underline{2.57} & \underline{2.59} \\
    RiLM \text{(Ours)} & \textbf{2.69} & \textbf{2.59} & \textbf{2.68} & \textbf{2.60} \\
\Xhline{2\arrayrulewidth}
\multicolumn{5}{l}{\scriptsize $^\dagger$continuously pre-trained language model}
\end{tabular}
}
}
\caption{BLEU score of \textproc{GrowOVER-Dialogue} on each month using Contriever. The table shows the BLEU score between the generated dialogue response and the gold dialogue response.} 
\label{tab:contrieverdialogue}
\end{center}
\end{table}

\clearpage
\newpage
\section{Algorithms}\label{generationAlgs}

\begin{algorithm}[H]
    \captionsetup{font=footnotesize}
    \caption{Initial Generation for \textproc{GrowOVER-QA}}\label{alg:initialQAGeneration}
    \scriptsize
    \textbf{Require:} $WP_{cur}$ := Wikipedia snapshots of the initial month\\
    $P$ := An empty array to store valid paragraphs\\
    $S$ := An empty array to store selected paragraphs\\
    $Q$ := An empty array to store generated QA pairs\\
    $\ast$ $article$ in $WP$ has attributes \textit{id}, \textit{title} and \textit{text}
    \medskip
    
    \begin{algorithmic}
    {
        \For{all $article$ $a_r \in WP_{init}$}
            \For{$paragraph$ $p_r \in a_r.text$}
                \If{$p_r$ is of adequate length}
                    \State $P$.append($p_r$)
                \EndIf
            \EndFor
            \State $S \leftarrow$ \textproc{ClusterParagraphs}($P$)
            \State $Q \leftarrow Q$ + \textproc{GenerateQA}($S$, \textproc{New})
        \EndFor
        
        \Statex
        \Function{ClusterParagraphs}{$P$}
            \State Extract features and obtain embeddings of $P$ using \textit{SimCSE}
            \State Compute cluster assignment using KMeans algorithm
            \State $S \leftarrow$ Randomly selected paragraphs from each cluster
            \State \Return {$S$}
        \EndFunction
        
        \Statex
        \Function{GenerateQA}{$S$, $type$}
            \State // initialize $QA$ as an empty array  
            \For{selected text $p \in S$}
                \State $qa \leftarrow$ generate QA pairs with $p$
                \State $qa.type \leftarrow type$
                \State $QA$.append($qa$)
            \EndFor
            \State \Return {$QA$}
        \EndFunction
    }

    \end{algorithmic}
\end{algorithm}

\begin{algorithm}[H]
    \captionsetup{font=footnotesize}
    \caption{Initial Generation for \textproc{GrowOVER-Dialogue}}\label{alg:initialDialogueGeneration}
    \scriptsize
    \textbf{Require:} $WP_{init}$ := Wikipedia snapshots at initial month\\ $P$ := An empty array to store splitted paragraphs\\
    $D$ := An empty array to store generated Dialogue\\ 
    $\ast$ \textit{article} in $WP$ has attributes \textit{id}, \textit{title} and \textit{text}
    \medskip
    
\begin{algorithmic}

\For{$article$ $a_r \in WP_{init}$}
    \State $P \leftarrow $ \textproc{SplitArticleIntoParagraph}($a_r$)
    \For{$paragraph$ $p \in P$}
        \If{$p$ is informative paragraph}
            \State {$D$.append(\textproc{GenerateDialogue}($p$, $a_r$.title))}
        \EndIf
    \EndFor
\EndFor

\Statex
\Function{SplitArticleIntoParagraph}{$a$}
    \For {$paragraph$ $p \in a.text$}
        \State {$P$.append(split $p$ into sentence)}
    \EndFor
    \State \Return {$P$}
\EndFunction

\Statex
\Function{GenerateDialogue}{$p$, $a$.title}
\State {$d \leftarrow$ generate dialogue with $p$ and $a$.title}
\For {$turn$ $t \in d$}
    \State {$t.type$ $\leftarrow$ \textproc{NEW}}
\EndFor
\State \Return d
\EndFunction
\end{algorithmic}
\end{algorithm}
\newpage

\newpage

\begin{algorithm}[H]
    \captionsetup{font=footnotesize}
    \caption{Temporal Updates for \textproc{GrowOVER-QA}}\label{alg:updateQA}
    \scriptsize
    \textbf{Require:} $W_{t}$ := Wikipedia snapshots of month $t$\\
    $QA_{t\text{-}1}$ := QAs of month $t$-1\\
    $QA_{t}$ := An empty set for QAs of month $t$\\
    \begin{algorithmic}
    {
        \For{$article$ $a_{t} \in W_{t}$}
            \For{$a_{t}$'s QA $qa \in QA_{t\text{-}1}$}
                \State $qa \leftarrow$ \textproc{UpdateQA}($qa$)
                \If{$qa.type$ is \textproc{Unchanged}}
                    \State $QA_{t}$.append($qa$)
                \EndIf
            \EndFor
            \State $G_{new} \leftarrow$ groups of \textit{new} sentences in $a_{t}$
            \For{group $g \in G_{new}$}
                \State $QA_{t}$.append(\textproc{GenerateQA}($g$, \textproc{New}))
            \EndFor
            \For{$changed$ sentence $s_{t} \in a_{t}$}
                \State $s_{t\text{-}1} \leftarrow$ contradictory sentence in $a_{t\text{-}1}$
                \State $QA_{t}$.append(\textproc{GenerateQA}($s_{t\text{-}1}$, $s_{t}$, \textproc{Changed}))
            \EndFor
        \EndFor

        \Statex
        \Function{UpdateQA}{$qa$}
            \If{all sentences $\in qa.evid\_text$ are \textit{unchanged}}
                \State $qa.type \leftarrow $ \textproc{Unchanged}
                \State $qa.index \leftarrow$ indices of $qa.evid\_text$ in $W_{t}$
            \Else
                \State $qa.type \leftarrow $ \textproc{Deleted}
            \EndIf
            \State \Return {$qa$}
        \EndFunction
    }

    \end{algorithmic}
\end{algorithm}

\begin{algorithm}[H]
    \captionsetup{font=footnotesize}
    \caption{Temporal Updates for \textproc{GrowOVER-Dialogue}}\label{alg:updateDialogue}
    \scriptsize
    \textbf{Require:} $W_{t}$ := Wikipedia snapshots on month $t$ \\
    $D_{t\text{-}1}$ := Dialogues of month $t$-1 \\
    $D_{t}$ := An empty set for dialogues of month $t$ \\

\begin{algorithmic}
\For {$article$ $a_{t} \in W_{t}$}
    \For {$a_{t}$'s $dialogue$ $d \in D_{t\text{-}1}$}
        \State {$d \leftarrow$ \textproc{UpdateDialog}($d$, $L_{t\text{-}1}$)}
        \State {$D_{t}$.append($d$)}
    \EndFor
    \State {$P \leftarrow$ \textproc{SplitArticleIntoParagraph}($a_{t}$)}
    \For {$paragraph$ $p \in P$}
        \If {$p$ is not informative paragraph} 
            \State
            \textbf{continue}
        \ElsIf {$\textit{changed}$ or $\textit{new}$ sentence in $p$}
            \State {$D_{t}$.append(\textproc{GenerateDial}($p$, $a_{t}$.title))}
        \EndIf
    \EndFor
\EndFor

\Statex
\Function{UpdateDialog}{$d$}
    \For {$turn$ $t \in d$}
        \If {$s$ is ${\textit{unchanged}}$ sentence}
            \State {$t.type$ $\leftarrow$ $\textproc{Unchanged}$}
            \State {$t.index$ $\leftarrow$ the index of $t.evid\_text$ in $WP_{cur}$} 
        \Else
            \State {$t.type$ $\leftarrow$ $\textproc{Deleted}$}
        \EndIf
    \EndFor
    \State \Return $d$
\EndFunction
\end{algorithmic}    

\end{algorithm}

\newpage
\section{Mturk Study}\label{app:mturk}

We thoroughly designed the sentence labeling process to accurately label sentences (e.g., setting a high threshold for selecting semantically identical sentences). To further validate using sentence similarity scores and natural language inference, we employed Amazon Mechanical Turk (AMT) workers to assess the sentence labels during the rebuttal process.

We randomly sampled 30 new and 30 changed sentences and asked AMT workers to classify whether each sentence in the new article was supported, not supported, or uncertain given the old article. For new sentences, the labeling is incorrect when classified as “supported,” while for changed sentences, the labeling is correct when classified as “not supported.” Each sentence was evaluated by three workers, and the majority vote was used. The human quality check results are shown in the table~\ref{tab:sentence_labeling}.

Although new sentences have lower accuracy since they are not verified with GPT-4, the results are still within acceptable bounds. Evaluating the correct label by humans requires a review of the entire article, which is extremely time-consuming. This becomes critical, especially for our dataset, which requires regular updates to reflect the ever-changing knowledge of the real world. Therefore, as evidenced by the human quality check results, our fully automated sentence labeling alone can efficiently provide reasonably accurate labels with no human effort at all.

\begin{table}[H]
\begin{center}
\renewcommand{\arraystretch}{1.2}
\resizebox{\linewidth}{!}{
{\footnotesize
\begin{tabular}{p{3.2cm} >{\centering\arraybackslash}p{1.8cm} >{\centering\arraybackslash}p{1.8cm}}
\Xhline{2\arrayrulewidth}
 & New & Changed \\
\hline
\textbf{\textproc{GrowOVER-QA}} \\ 
\hline
    Accuracy & 86.7 & 96.7 \\
\Xhline{2\arrayrulewidth}
\end{tabular}
}
}
\caption{Sentence labeling validation.}
\label{tab:sentence_labeling}
\end{center}
\end{table}

\newpage
\section{Data Analysis}\label{dataAnalysis}
\begin{table}[H]
\begin{center}
\footnotesize
\renewcommand{\arraystretch}{1.2}
\resizebox{\linewidth}{!}{
\begin{tabular}{c | c c c c c}
\Xhline{2\arrayrulewidth}
 & Unchanged & New & Changed & Deleted & Total\\ 
\hline
08 & - & 32,807 & - & - & 32,807\\
09 & 32,422 (512) & 4,936 (512) & 290 (245) & 385 & 37,648 (1,269)\\
10 & 36,863 & 5,193 & 307 & 785 & 42,363\\ 
11 & 41,257 & 5,363 & 309 & 1,106 & 46,929\\ 
12 & 43,422 & 5,082 & 313 & 1,211 & 48,817\\ 
\Xhline{2\arrayrulewidth}
\end{tabular}
}
\caption{QA}
\label{tab:qaStatistics}
\end{center}
\end{table}

\vspace{-0.5cm}
\begin{table}[H]
\begin{center}
\small
\renewcommand{\arraystretch}{1.2}
\resizebox{\linewidth}{!}{
\begin{tabular}{c | c c c c c}
\Xhline{2\arrayrulewidth}
 & Unchanged & New & Changed & Deleted & Total\\ 
\hline
08 & - & 108,128 & - & - & 108,128 \\ 
09 & 109,752 (512) & 4,478 (512) & 156 (133) & 987 & 114,386 (1,157)\\
10 & 115,022 & 4,797 & 147 & 1,883 & 119,966 \\
11 & 120,551 & 4,870 & 158 & 2,161 & 125,579 \\
12 & 125,940 & 5,218 & 142 & 2,427 & 131,300 \\
\Xhline{2\arrayrulewidth}
\end{tabular}
}
\caption{Dialogue}
\label{tab:dialogueStatistics}
\end{center}
\end{table}
\begin{figure*}[t]
    \centering
    \includegraphics[width=\textwidth]{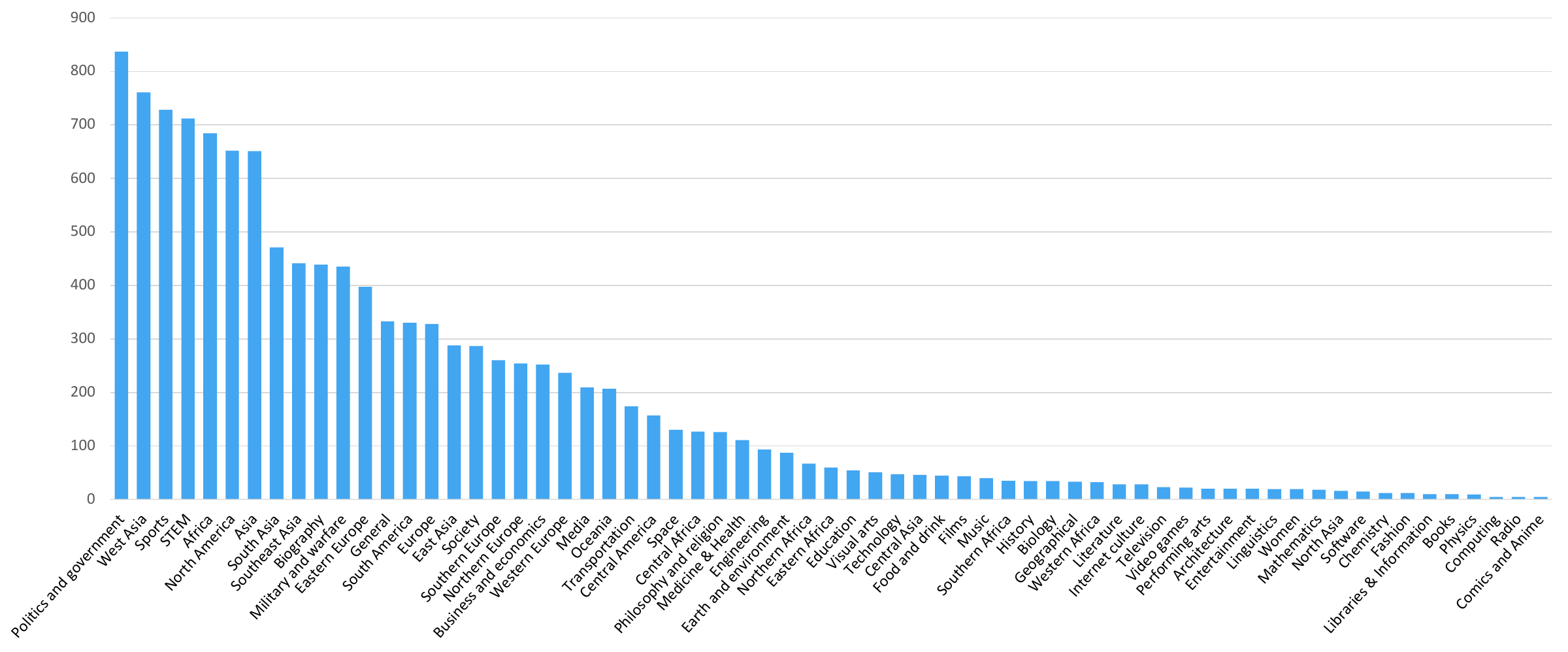}
    \caption{Article Categories Overview}
    \label{fig:article_categories}
\end{figure*}

\section{Dataset Examples}\label{datasetExamples}
Table~\ref{tab:anExampleOfQA} shows the examples of QA.
Table~\ref{tab:anExampleOfDialogue} and~\ref{tab:anotherExampleOfDialogue} show the examples of Dialogue.

\newpage
\begin{table*}
\small
\renewcommand{\arraystretch}{1.2}
\resizebox{\textwidth}{!}{
\begin{tabular}{p{\linewidth}}
    \Xhline{3\arrayrulewidth}
        \textbf{Article :} Politics of Cambodia\\
        \hline
        \textbf{Type :} \textproc{Changed}\\
        \textbf{Question}: Who is the current prime minister of Cambodia from the Cambodian People's Party (CPP)? \\
        \textbf{Answer}: Hun Manet \\
        \textbf{Previous Answer}: Hun Sen \\
        \textbf{Evidence Text}: The current prime minister is Cambodian People's Party (CPP) member Hun Manet. \\
        \textbf{Evidence Index}: 80 \\
    \Xhline{2\arrayrulewidth}
        \textbf{Article :} The Eras Tour\\
        \hline
        \textbf{Type :} \textproc{New}\\
        \textbf{Question}: Who is directing the concert film \textit{Taylor Swift: The Eras Tour}? \\
        \textbf{Answer}: Sam Wrench \\
        \textbf{Evidence Text}: On August 31, 2023, Swift announced the concert film \textit{Taylor Swift: The Eras Tour}, directed by Sam Wrench. Recorded at SoFi Stadium in Los Angeles, the film is scheduled for release to theaters in North America on October 13. \\
        \textbf{Evidence Index}: 263 \\
    \Xhline{2\arrayrulewidth}
        \textbf{Article :} Airline\\
        \hline
        \textbf{Type :} \textproc{Unchanged}\\
        \textbf{Question}: Which airport would most Manhattan travelers prefer for its proximity? \\
        \textbf{Answer}: LaGuardia Airport \\
        \textbf{Evidence Text}: For example, LaGuardia Airport is the preferred airport for most of Manhattan due to its proximity, while long-distance routes must use John F. Kennedy International Airport's longer runways. \\
        \textbf{Evidence Index}: 297 \\
        \Xhline{3\arrayrulewidth}
\end{tabular}
}
\caption{An Example of QA}
\label{tab:anExampleOfQA}
\end{table*}

\newpage
\begin{table*}
\small
\renewcommand{\arraystretch}{1.2}
\resizebox{\textwidth}{!}{
\begin{tabular}{p{\linewidth}}
    \Xhline{3\arrayrulewidth}
        \textbf{Article :} 2030 Winter Olympics \\
        \hline
        \textbf{Paragraph :}\\
        $^{20}$A decision on the 2030 Winter Olympics host city has been delayed until 23 July 2024, to allow the IOC more time to carefully plan the future of the Winter Olympics. $^{21}$Due to impacts from climate change, the IOC is considering a number of changes to future games, such as rotating host cities, limiting the games to existing or temporary venues, and establishing minimum average temperature requirements. $^{22}$It also considered awarding the 2030 and 2034 winter games simultaneously at the next IOC session, but IOC President Thomas Bach has ruled out that possibility because "it would not be the right thing to do". $^{23}$However, at the 141st IOC Session in Mumbai, where these games were originally set to be awarded, the IOC approved to award both the 2030 and 2034 Winter Olympics at the 142nd IOC Session in Paris on the eve of the 2024 Summer Olympics. \\
        \hline
        \textbf{Dialogue :}\\
        \arrayrulecolor{lightgray}\hline
        \textbf{Type}: \textproc{Unchanged} \\
        \textbf{User}: I was looking forward to finding out which city will host the 2030 Winter Olympics, but I heard there's been a delay. What's going on with that? \\
        \textbf{Expert}: Yes, that's correct. The decision has been postponed until 23 July 2024. The International Olympic Committee (IOC) wants to take more time to plan the future of the Winter Games with careful consideration. \\
        \textbf{Evidence Index}: 20 \\
        \arrayrulecolor{lightgray}\hline
        \textbf{Type}: \textproc{Unchanged} \\ 
        \textbf{User}: I guess the delay must be frustrating for the candidate cities. I've heard that the changing climate is causing some issues? \\
        \textbf{Expert}: Indeed, the changing climate is a significant concern. The IOC is exploring various adjustments for future games, including the possibility of having rotating host cities and using existing or temporary venues to mitigate the impact. \\
        \textbf{Evidence Index}: 21 \\
        \arrayrulecolor{lightgray}\hline
        \textbf{Type}: \textproc{Unchanged} \\ 
        \textbf{User}: I also caught a rumor that they might announce the hosts for both 2030 and 2034 at the same time. Is that true? \\
        \textbf{Expert}: That was a consideration at one point. However, IOC President Thomas Bach has stated that awarding both the 2030 and 2034 Winter Olympics simultaneously wouldn't be appropriate. \\
        \textbf{Evidence Index}: 22 \\
        \arrayrulecolor{lightgray}\hline
        \textbf{Type}: \textproc{New} \\ 
        \textbf{User}: So, when can we expect the announcement for the 2030 host city? \\
        \textbf{Expert}: The announcement is set to be made at the 142nd IOC Session in Paris, which will happen right before the 2024 Summer Olympics. They will also award the 2034 Winter Olympics at that session. \\
        \textbf{Evidence Index}: 23 \\
        \Xhline{3\arrayrulewidth}
\end{tabular}
}
\caption{An Example of Dialogue}
\label{tab:anExampleOfDialogue}
\end{table*}

\begin{table*}
\small
\renewcommand{\arraystretch}{1.2}
\resizebox{\textwidth}{!}{
\begin{tabular}{p{\linewidth}}
    \Xhline{3\arrayrulewidth}
        \textbf{Article :} Tornadoes of 2023 \\
        \hline
        \textbf{Paragraph :}\\
        $^{357}$More rounds of severe storms occurred on June 17, including a powerful MCS that pushed through Northern and Central Oklahoma. $^{358}$Both the Oklahoma City and Tulsa metropolitan areas had widespread wind damage with the latter metropolitan area seeing wind gusts of up to and multiple weak tornadoes. $^{359}$On June 18, more tornadoes were confirmed in Florida and Mississippi, including an EF3 tornado which caused significant damage in Louin, Mississippi, where at least one fatality and 20 injuries occurred. $^{360}$Scattered tornadoes occurred across the Southeast on June 19 as well, including an EF2 tornado that injured six people in Moss Point, Mississippi. $^{361}$Overall, 87 tornadoes were confirmed during this outbreak sequence.\\
        \hline
        \textbf{Dialogue :}\\
        \arrayrulecolor{lightgray}\hline
        \textbf{Type}: \textproc{Unchanged} \\ 
        \textbf{User}: I heard there were some intense storms in Oklahoma last June. What happened exactly? \\
        \textbf{Expert}: Indeed, on June 17, Oklahoma experienced severe storms, including a particularly strong Mesoscale Convective System that affected the northern and central parts of the state. \\
        \textbf{Evidence Index}: 357 \\
        \arrayrulecolor{lightgray}\hline
        \textbf{Type}: \textproc{Unchanged} \\ 
        \textbf{User}: That sounds serious. Did the storms cause a lot of damage in the area? \\
        \textbf{Expert}: Yes, the storms led to extensive wind damage, especially in the Oklahoma City and Tulsa areas. Tulsa even experienced wind gusts of up to 80 mph and several weak tornadoes. \\
        \textbf{Evidence Index}: 358 \\
        \arrayrulecolor{lightgray}\hline
        \textbf{Type}: \textproc{Unchanged} \\ 
        \textbf{User}: I also caught wind of some tornado activity in other states around that time. \\ 
        \textbf{Expert}: That's correct. The following day, on June 18, Florida and Mississippi reported tornadoes. Mississippi was hit by an EF3 tornado in Louin, which resulted in one death and 20 injuries.\\
        \textbf{Evidence Index}: 359 \\
        \arrayrulecolor{lightgray}\hline
        \textbf{Type}: \textproc{Unchanged} \\ 
        \textbf{User}: Was that the end of it, or did the tornadoes continue? \\ 
        \textbf{Expert}: The tornadoes continued. On June 19, the Southeast saw more, including an EF2 tornado in Moss Point, Mississippi, which injured six people.\\
        \textbf{Evidence Index}: 360 \\
        \arrayrulecolor{lightgray}\hline
        \textbf{Type}: \textproc{Changed} \\ 
        \textbf{User}: It must have been a large outbreak to affect so many areas. \\ 
        \textbf{Expert}: It was indeed a significant outbreak. In total, there were 87 confirmed tornadoes during that sequence of storms. \\ 
        \textbf{Evidence Index}: 361 \\
        \Xhline{3\arrayrulewidth}
\end{tabular}
}
\caption{Another Example of Dialogue}
\label{tab:anotherExampleOfDialogue}
\end{table*}

\section{Case Study}\label{caseStudy}
Table~\ref{tab:CaseStudyQA} and Table~\ref{tab:CaseStudyQA2} present a case study from \textproc{GrowOVER-QA}. \textbf{Predictions} denote the answers generated by each retriever-generator framework. Table~\ref{tab:CaseStudyDialogue} presents a case study from \textproc{GrowOVER-Dialogue}.
\begin{table*}
\small
\renewcommand{\arraystretch}{1.2}
\resizebox{\textwidth}{!}{
\begin{tabular}{p{\linewidth}}
    \Xhline{3\arrayrulewidth}
        \textbf{Article :} Football player\\
        \hline
        \textbf{Type :} \textproc{Unchanged}\\
        \textbf{Question}: What was the average annual salary for goalkeepers in Major League Soccer during the 2013 season? \\
        \textbf{Answer}: \$85,296 \\
        \textbf{Evidence Text}: For example, the average annual salary for footballers in Major League Soccer for the 2013 season was \$148,693, with significant variations depending on the player position (goalkeepers for example earned \$85,296, whereas forwards earned \$251,805. \\
        \hline
        \textbf{Predictions :} \\
        \textbf{Vanilla}: \$110,000 \;
        \textbf{RaLM}: \$148,693 \;
        \textbf{RaLM-CP}: \$148,693 \;
        \textbf{RaLM-$D^{*}$ \text{(Ours)}}: \$85,296 \;
        \textbf{RiLM \text{(Ours)}}: \$85,296 \\
        \hline
        \textbf{Retrieved Documents :} \\
        \textbf{[Top-3]} \\
        (1) \textbf{\textit{Hit}}: However, only a fraction of men's professional football players is paid at this level. ... \textbf{For example, the average annual salary for footballers in Major League Soccer for the 2013 season was \$148,693, with significant variations depending on the player position (goalkeepers for example earned \$85,296, whereas forwards earned \$251,805.} Popularity and average salaries in women's leagues are far lower. For example, players in ... \\
        (2) \textit{Miss}: MLS has a set of pool goalkeepers who are signed to a contract with the league and are loaned to teams during emergencies in which they are missing a goalkeeper due to injuries or suspensions. ... These initiatives have brought about an increase in on-field competition. \\
        (3) \textit{Miss}: According to "France Football", Messi was the world's highest-paid footballer for five years out of six between 2009 and 2014; ... In 2020, Messi became the second footballer, as well as the second athlete in a team sport, after Cristiano Ronaldo, to surpass \$1 billion in earnings during their careers. \\ \\
        \textbf{[RaLM-$D^{*}$]} \\
        \textbf{\textit{Hit}}: However, only a fraction of men's professional football players is paid at this level. ... \textbf{For example, the average annual salary for footballers in Major League Soccer for the 2013 season was \$148,693, with significant variations depending on the player position (goalkeepers for example earned \$85,296, whereas forwards earned \$251,805.} Popularity and average salaries in women's leagues are far lower. For example, players in ... \\ \\
        \textbf{[RiLM]} \\
        \textbf{\textit{Hit}}: However, only a fraction of men's professional football players is paid at this level. ... \textbf{For example, the average annual salary for footballers in Major League Soccer for the 2013 season was \$148,693, with significant variations depending on the player position (goalkeepers for example earned \$85,296, whereas forwards earned \$251,805.} Popularity and average salaries in women's leagues are far lower. For example, players in ... \\
    \Xhline{3\arrayrulewidth}
        \textbf{Article :} Benjamin Netanyahu \\
        \hline
        \textbf{Type :} \textproc{New}\\
        \textbf{Question}: What city was Benjamin Netanyahu born in? \\
        \textbf{Answer}: Tel Aviv \\
        \textbf{Evidence Text}: Netanyahu was born in Tel Aviv, to Benzion Netanyahu (original name Mileikowsky) and Tzila (Cela; \\
        \hline
        \textbf{Predictions :} \\
        \textbf{Vanilla}: Jerusalem, Israel \;
        \textbf{RaLM}: Tel Aviv \;
        \textbf{RaLM-CP}: Tel Aviv \;
        \textbf{RaLM-$D^{*}$ \text{(Ours)}}: Tel Aviv \;
        \textbf{RiLM \text{(Ours)}}: Tel Aviv \\
        \hline
        \textbf{Retrieved Documents :} \\
        \textbf{[Top-3]} \\
        (1) \textit{Miss}: Netanyahu was the second of three children. He was initially raised and educated in Jerusalem, where he attended ... the liberal sensibilities of the Reform synagogue, Temple Judea of Philadelphia, that the family attended. \\
        (2) \textbf{\textit{Hit}}: \textbf{Netanyahu was born in Tel Aviv, to Benzion Netanyahu (original name Mileikowsky) and Tzila (Cela;} née Segal). His mother was born in 1912 in Petah Tikva, then in Ottoman Palestine, now Israel. Though all his grandparents were born in ... \\
        (3) \textit{Miss}: Netanyahu made his closeness to Donald Trump, a personal friend since the 1980s, central to his political appeal in Israel from 2016. ... He claims descent from the Vilna Gaon. \\ \\
        \textbf{[RaLM-$D^{*}$]} \\
        \textbf{\textit{Hit}}: \textbf{Netanyahu was born in Tel Aviv, to Benzion Netanyahu (original name Mileikowsky) and Tzila (Cela;} née Segal). His mother was born in 1912 in Petah Tikva, then in Ottoman Palestine, now Israel. Though all his grandparents were born in ... \\ \\
        \textbf{[RiLM]} \\
        \textbf{\textit{Hit}}: \textbf{Netanyahu was born in Tel Aviv, to Benzion Netanyahu (original name Mileikowsky) and Tzila (Cela;} née Segal). His mother was born in 1912 in Petah Tikva, then in Ottoman Palestine, now Israel. Though all his grandparents were born in ... \\
        \Xhline{3\arrayrulewidth}
\end{tabular}
}
\caption{Case Study for QA}
\label{tab:CaseStudyQA}
\end{table*}

\begin{table*}
\small
\renewcommand{\arraystretch}{1.2}
\resizebox{\textwidth}{!}{
\begin{tabular}{p{\linewidth}}
    \Xhline{3\arrayrulewidth}
        \textbf{Article :} Kyrylo Budanov\\
        \hline
        \textbf{Type :} \textproc{Changed}\\
        \textbf{Question}: What is Kyrylo Budanov's military rank? \\
        \textbf{Answer}: Lieutenant general \\
        \textbf{Evidence Text}: He holds the rank of lieutenant general. \\
        \hline
        \textbf{Predictions :} \\
        \textbf{Vanilla}: Kyrylo Budanov is a Major General in the Ukrainian Armed Forces. \;
        \textbf{RaLM}: Lieutenant General \;
        \textbf{RaLM-CP}: Lieutenant General \;
        \textbf{RaLM-$D^{*}$ \text{(Ours)}}: Lieutenant General \;
        \textbf{RiLM \text{(Ours)}}: Lieutenant General \\
        \hline
        \textbf{Retrieved Documents :} \\
        \textbf{[Top-3]} \\
        (1) \textbf{\textit{Hit}}: Kyrylo Oleksiyovych Budanov (; born 4 January 1986) is a Ukrainian military leader who is the chief of the ... Budanov previously served as the Deputy Director of one of the Departments of the Foreign Intelligence Service of Ukraine. \textbf{He holds the rank of lieutenant general.} ... as head of the Main Intelligence Directorate of the Ministry of Defense. \\
        (2) \textit{Miss}: The Austrian military has a wide variety of equipment. Recently, Austria has spent considerable amounts of money modernizing its military arsenal. ... \\
        (3) \textit{Miss}: Soon after the start of the German invasion of the Soviet Union, he was soon re-drafted into the Red Army on 4 July 1941 and initially deployed to the front as part of the 50th Cavalry Regiment. ... \\ \\
        \textbf{[RaLM-$D^{*}$]} \\
        \textbf{\textit{Hit}}: Kyrylo Oleksiyovych Budanov (; born 4 January 1986) is a Ukrainian military leader who is the chief of the ... Budanov previously served as the Deputy Director of one of the Departments of the Foreign Intelligence Service of Ukraine. \textbf{He holds the rank of lieutenant general.} ... as head of the Main Intelligence Directorate of the Ministry of Defense. \\ \\
        \textbf{[RiLM]} \\
        \textbf{\textit{Hit}}: Kyrylo Oleksiyovych Budanov (; born 4 January 1986) is a Ukrainian military leader who is the chief of the ... Budanov previously served as the Deputy Director of one of the Departments of the Foreign Intelligence Service of Ukraine. \textbf{He holds the rank of lieutenant general.} ... as head of the Main Intelligence Directorate of the Ministry of Defense. \\
    \Xhline{3\arrayrulewidth}
        \textbf{Article :} Darwin, Northern Territory \\
        \hline
        \textbf{Type :} \textproc{Unchanged}\\
        \textbf{Question}: What is the name of the passenger train service that connects Darwin to Adelaide? \\
        \textbf{Answer}: The Ghan \\
        \textbf{Evidence Text}: The first service ran in 2004. "The Ghan" passenger train service from Adelaide via Alice Springs and Katherine runs once per week in each direction, with some exceptions. \\
        \hline
        \textbf{Predictions :} \\
        \textbf{Vanilla}: The Ghan \;
        \textbf{RaLM}: The Ghan \;
        \textbf{RaLM-CP}: The Ghan \;
        \textbf{RaLM-$D^{*}$ \text{(Ours)}}: The Ghan \;
        \textbf{RiLM \text{(Ours)}}: The Ghan \\
        \hline
        \textbf{Retrieved Documents :} \\
        \textbf{[Top-3]} \\
        (1) \textbf{\textit{Hit}}: The Territory's public transport services are managed by the Department of Lands and Planning, ... \textbf{The first service ran in 2004. "The Ghan" passenger train service from Adelaide via Alice Springs and Katherine runs once per week in each direction, with some exceptions.} Historically, ... \\
        (2) \textit{Miss}: Darwin and its suburbs, Palmerston and Katherine, are powered by the Channel Island Power Station, the Northern Territory's largest power plant, and the ... \\
        (3) \textit{Miss}: The platform layout at Flinders Street is unusual among Australian terminal stations for being almost entirely composed of through tracksa product ... \\ \\
        \textbf{[RaLM-$D^{*}$]} \\
        \textbf{\textit{Hit}}: The Territory's public transport services are managed by the Department of Lands and Planning, ... \textbf{The first service ran in 2004. "The Ghan" passenger train service from Adelaide via Alice Springs and Katherine runs once per week in each direction, with some exceptions.} Historically, ... \\ \\
        \textbf{[RiLM]} \\
        \textbf{\textit{Hit}}: The Territory's public transport services are managed by the Department of Lands and Planning, ... \textbf{The first service ran in 2004. "The Ghan" passenger train service from Adelaide via Alice Springs and Katherine runs once per week in each direction, with some exceptions.} Historically, ... \\
        \Xhline{3\arrayrulewidth}
\end{tabular}
}
\caption{Another Case Study for QA}
\label{tab:CaseStudyQA2}
\end{table*}


\begin{table*}
\small
\renewcommand{\arraystretch}{1.2}
\resizebox{\textwidth}{!}{
\begin{tabular}{p{\linewidth}}
    \Xhline{3\arrayrulewidth}
    \textbf{Article :} Marshall Islands\\
    \Xhline{3\arrayrulewidth}
        \textbf{Turn 1}\\
        \textbf{Type :} \textproc{Changed}\\
        \textbf{User}: I've heard that the Marshall Islands were inhabited for a very long time. What kind of influence did the early settlers have on these islands? \\
        \textbf{Expert}: Indeed, the islands have been inhabited since the 2nd millennium BC. The early Austronesian settlers brought with them Southeast Asian crops like coconuts and breadfruit, as well as domesticated chickens, which were crucial for making the islands habitable. \\
        \textbf{Evidence Text}: Austronesian settlers reached the Marshall Islands as early as the 2nd millennium BC and introduced Southeast Asian crops, including coconuts, giant swamp taro, and breadfruit, as well as domesticated chickens which made the islands permanently habitable. \\
        \hline
        \textbf{Predictions :} \\
        \textbf{Vanilla}: The Marshall Islands were inhabited for a very long time. The early settlers had a big influence on the islands. They built houses, farmed, and fished. They also made tools and weapons. The early settlers also had a big influence on the culture of the Marshall Islands. They taught the people how to make clothes, how to cook food, and how to make music. The early settlers also taught the people how to make weapons and how to fight. The early settlers \\
        \textbf{RaLM}: The early settlers of the Marshall Islands introduced Southeast Asian crops, including coconuts, giant swamp taro, and breadfruit, as well as domesticated chickens which made the islands permanently habitable. They developed a complex technique for interisland piloting by reading disruptions in ocean swells caused by low-lying coral atolls. \\
        \textbf{RaLM-CP}: The early settlers of the Marshall Islands introduced Southeast Asian crops, including coconuts, giant swamp taro, and breadfruit, as well as domesticated chickens which made the islands permanently habitable. They developed a complex technique for interisland piloting by reading disruptions in ocean swells caused by low-lying coral atolls. \\
        \textbf{RaLM-$D^{*}$ \text{(Ours)}}: The early settlers introduced Southeast Asian crops, including coconuts, giant swamp taro, and breadfruit, as well as domesticated chickens which made the islands permanently habitable. \\
        \textbf{RiLM \text{(Ours)}}: The early settlers introduced Southeast Asian crops, including coconuts, giant swamp taro, and breadfruit, as well as domesticated chickens which made the islands permanently habitable. \\
        \hline
        \textbf{Retrieved Documents :} \\
        \textbf{[Top-3]} \\
        (1) \textit{Miss}: The majority of the citizens of the Republic of Marshall Islands are of Marshallese descent, ... around the 1st century CE at the village of Laura on Majuro and on Kwajalein Atoll. \\
        (2) \textit{\textbf{Hit}}: \textbf{Austronesian settlers reached the Marshall Islands as early as the 2nd millennium BC and introduced Southeast Asian crops, including coconuts, giant swamp taro, and breadfruit, as well as domesticated chickens which made the islands permanently habitable.} They developed a complex technique for ... \\
        (3) \textit{Miss}: The British sea captains John Marshall and Thomas Gilbert visited the islands in 1788. Their vessels had been part of the First Fleet taking convicts from England to Botany Bay ... \\ \\
        \textbf{[RaLM-$D^{*}$]} \\
        \textit{\textbf{Hit}}: \textbf{Austronesian settlers reached the Marshall Islands as early as the 2nd millennium BC and introduced Southeast Asian crops, including coconuts, giant swamp taro, and breadfruit, as well as domesticated chickens which made the islands permanently habitable.} They developed a complex technique for ... \\ \\
        \textbf{[RiLM]} \\
        \textit{\textbf{Hit}}: \textbf{Austronesian settlers reached the Marshall Islands as early as the 2nd millennium BC and introduced Southeast Asian crops, including coconuts, giant swamp taro, and breadfruit, as well as domesticated chickens which made the islands permanently habitable.} They developed a complex technique for ... \\
    \Xhline{3\arrayrulewidth}
        \textbf{Turn 2}\\
        \textbf{Type :} \textproc{New}\\
        \textbf{User}: That's fascinating. I wonder how they navigated between the islands back then. \\
        \textbf{Expert}: They actually developed a sophisticated method for navigating the ocean. They could read the disruptions in ocean swells, which were caused by the coral atolls, to pilot between the islands. \\
        \textbf{Evidence Text}: They developed a complex technique for interisland piloting by reading disruptions in ocean swells caused by low-lying coral atolls. \\
        \hline
        \textbf{Predictions :} \\
        \textbf{Vanilla}: The early settlers developed a complex technique for interisland piloting by reading disruptions in ocean swells caused by low-lying coral atolls. \\
        \textbf{RaLM}: The early settlers developed a complex technique for interisland piloting by reading disruptions in ocean swells caused by low-lying coral atolls. \\
        \textbf{RaLM-CP}: The early settlers developed a complex technique for interisland piloting by reading disruptions in ocean swells caused by low-lying coral atolls. \\
        \textbf{RaLM-$D^{*}$ \text{(Ours)}}: The early settlers developed a complex technique for interisland piloting by reading disruptions in ocean swells caused by low-lying coral atolls. \\
        \textbf{RiLM \text{(Ours)}}: The early settlers developed a complex technique for interisland piloting by reading disruptions in ocean swells caused by low-lying coral atolls. \\
        \hline
\end{tabular}
}
\end{table*}

\begin{table*}
\small
\renewcommand{\arraystretch}{1.2}
\resizebox{\textwidth}{!}{
\begin{tabular}{p{\linewidth}}
    \hline
        \textbf{Retrieved Documents :} \\
        \textbf{[Top-3]} \\
        (1) \textit{\textbf{Hit}}: ... which made the islands permanently habitable. \textbf{They developed a complex technique for interisland piloting by reading disruptions in ocean swells caused by low-lying coral atolls.} In 1526, Spanish explorer Alonso de Salazar ... \\
        (2) \textit{Miss}: The majority of the citizens of the Republic of Marshall Islands are of Marshallese descent, ... around the 1st century CE at the village of Laura on Majuro and on Kwajalein Atoll. \\
        (3) \textit{Miss}: The Austronesian settlers introduced Southeast Asian crops, including coconuts, giant swamp taro, and breadfruit, as well as domesticated chickens throughout the Marshall Islands. ... \\ \\
        \textbf{[RaLM-$D^{*}$]} \\
        \textit{\textbf{Hit}}: ... which made the islands permanently habitable. \textbf{They developed a complex technique for interisland piloting by reading disruptions in ocean swells caused by low-lying coral atolls.} In 1526, Spanish explorer Alonso de Salazar ... \\ \\
        \textbf{[RiLM]} \\
        \textit{\textbf{Hit}}: ... which made the islands permanently habitable. \textbf{They developed a complex technique for interisland piloting by reading disruptions in ocean swells caused by low-lying coral atolls.} In 1526, Spanish explorer Alonso de Salazar ... \\
        \Xhline{3\arrayrulewidth}
        \textbf{Turn 3}\\
        \textbf{Type :} \textproc{Unchanged}\\
        \textbf{User}: I guess that means they were pretty isolated until Europeans arrived. \\
        \textbf{Expert}: That's correct. The first European who might have sighted the islands was Spanish explorer Alonso de Salazar in 1526, and the first confirmed European landing was by Álvaro de Saavedra Cerón two years later on an uninhabited island. \\
        \textbf{Evidence Text}: In 1526, Spanish explorer Alonso de Salazar may have been the first European to sight the islands, and Álvaro de Saavedra Cerón made landfall on an uninhabited island two years later. \\
        \hline
        \textbf{Predictions :} \\
        \textbf{Vanilla}: That's correct. The Marshall Islands were not discovered by Europeans until 1529, when Spanish explorer Alvaro de Mendana de Neira arrived. He named the islands after his patron, the Spanish admiral Álvaro de Mendaña de Neyra. \\
        \textbf{RaLM}: That's right. The islands were largely untouched by European explorers until the late 16th century. \\
        \textbf{RaLM-CP}: That's right. The islands were largely untouched by European explorers until the late 16th century. \\
        \textbf{RaLM-$D^{*}$ \text{(Ours)}}: That's correct. The islands were largely untouched by European explorers until the 18th century. \\
        \textbf{RiLM \text{(Ours)}}: That's correct. The islands were largely untouched by European explorers until the 18th century. \\
        \hline
        \textbf{Retrieved Documents :} \\
        \textbf{[Top-3]} \\
        (1) \textit{\textbf{Hit}}: ... by low-lying coral atolls. \textbf{In 1526, Spanish explorer Alonso de Salazar may have been the first European to sight the islands, and Álvaro de Saavedra Cerón made landfall on an uninhabited island two years later.} Several other Spanish ... \\
        (2) \textit{Miss}: The majority of the citizens of the Republic of Marshall Islands are of Marshallese descent, ... around the 1st century CE at the village of Laura on Majuro and on Kwajalein Atoll. \\
        (3) \textit{Miss}: The British sea captains John Marshall and Thomas Gilbert visited the islands in 1788. Their vessels had been part of the First Fleet taking convicts from England to Botany Bay in New South Wales, ... \\ \\
        \textbf{[RaLM-$D^{*}$]} \\
        \textit{\textbf{Hit}}: ... by low-lying coral atolls. \textbf{In 1526, Spanish explorer Alonso de Salazar may have been the first European to sight the islands, and Álvaro de Saavedra Cerón made landfall on an uninhabited island two years later.} Several other Spanish ... \\ \\
        \textbf{[RiLM]} \\
        \textit{\textbf{Hit}}: ... by low-lying coral atolls. \textbf{In 1526, Spanish explorer Alonso de Salazar may have been the first European to sight the islands, and Álvaro de Saavedra Cerón made landfall on an uninhabited island two years later.} Several other Spanish ... \\
    \Xhline{3\arrayrulewidth}
\end{tabular}
}
\end{table*}

\begin{table*}
\small
\renewcommand{\arraystretch}{1.2}
\resizebox{\textwidth}{!}{
\begin{tabular}{p{\linewidth}}
    \Xhline{3\arrayrulewidth}
        \textbf{Turn 4}\\
        \textbf{Type :} \textproc{Unchanged}\\
        \textbf{User}: And how did the Marshall Islands get their name? \\
        \textbf{Expert}: The name comes from the British captains John Marshall and Thomas Gilbert who landed on Mili Atoll in 1788. It was the first recorded European presence in over 200 years, and subsequent maps began to refer to the islands by Captain Marshall's name. \\
        \textbf{Evidence Text}: Later maps and charts named the islands after Captain Marshall. \\
        \hline
        \textbf{Predictions :} \\
        \textbf{Vanilla}: The islands were named after British explorer John Marshall, who was the first European to map them in 1788. \\
        \textbf{RaLM}: The islands were named after Captain John Marshall, who was the first European to sight them. \\
        \textbf{RaLM-CP}: The islands were named after Captain John Marshall, who was the first European to sight them. \\
        \textbf{RaLM-$D^{*}$ \text{(Ours)}}: The islands were named after Captain John Marshall, who was the first European to map the islands in 1788. \\
        \textbf{RiLM \text{(Ours)}}: The islands were named after Captain John Marshall, who was the first European to map the islands in 1788. \\
        \hline
        \textbf{Retrieved Documents :} \\
        \textbf{[Top-3]} \\
        (1) \textit{\textbf{Hit}}: ... in the archipelago in over 200 years. \textbf{Later maps and charts named the islands after Captain Marshall.} ... \\
        (2) \textit{Miss}: The majority of the citizens of the Republic of Marshall Islands are of Marshallese descent, ... around the 1st century CE at the village of Laura on Majuro and on Kwajalein Atoll. \\
        (3) \textit{Miss}: The British sea captains John Marshall and Thomas Gilbert visited the islands in 1788. Their vessels had been part of the First Fleet taking convicts from England to Botany Bay in New South Wales, ... \\ \\
        \textbf{[RaLM-$D^{*}$]} \\
        \textit{\textbf{Hit}}: ... in the archipelago in over 200 years. \textbf{Later maps and charts named the islands after Captain Marshall.} ... \\ \\
        \textbf{[RiLM]} \\
        \textit{\textbf{Hit}}: ... in the archipelago in over 200 years. \textbf{Later maps and charts named the islands after Captain Marshall.} ... \\
    \Xhline{3\arrayrulewidth}
\end{tabular}
}
\caption{Case Study for Dialogue}
\label{tab:CaseStudyDialogue}
\end{table*}
\section{Prompt examples}\label{ex:prompt}
Table~\ref{tab:initialQAPrompt} shows the prompt used for the initial generation of the QA pair. Table ~\ref{tab:newQAPrompt} and ~\ref{tab:newQAPrompt2} show the prompts used to generate \textproc{New} QA pairs, without and with Source Content respectively. Table~\ref{tab:changedQAPrompt} shows the prompt used to generate \textproc{Changed} QA pairs.
Table~\ref{tab:generateDialoguePrompt} shows the prompt used for Dialogue generation.

\definecolor{Gray}{gray}{0.93}
\begin{table*}
\small
\renewcommand{\arraystretch}{1.4}
\rowcolors{1}{Gray}{Gray}
\resizebox{\textwidth}{!}{
\begin{tabular}{!{\vrule width 1.2pt} p{\linewidth} !{\vrule width 1.2pt}}
    \Xhline{1.2pt}
        Generate a Q\&A pair based on a given context, where the context is understood but NOT DIRECTLY VISIBLE to the person answering the question. Assume the person answering the question has common sense and is aware of the details and key points in the paragraph, but the paragraph itself is not quoted or referenced directly.\\
        \\
        Paragraph (a list of sentences): \{\textit{paragraph}\}\\
        \\
        Use the following instructions for generating a Q\&A pair:\\
        1) Provide a question, an answer, and a bounding box.\\
        2) DON’T use phrases such as ‘according to the paragraph’ in your question.\\
        3) An answer should be an entity or entities. Provide a SHORT ANSWER.\\
        4) The bounding box for a paragraph is defined as (starting sentence index, ending sentence index): the bounding box should be sufficiently large to encompass all the information necessary for a reader to FULLY infer the answer to the question.\\
        5) The sentence index starts from 0.\\
        6) Generate a SINGLE Q\&A pair.\\
        \\
        Be sure to follow the following format and write your answer within curly brackets.\\
        The format is as follows:\\
        \{Question\}\{Answer\}\{starting sentence index\}\{ending sentence index\}\\
    \Xhline{1.2pt}
\end{tabular}
}

\caption{Sample prompt for initial generation of a QA pair}
\label{tab:initialQAPrompt}
\end{table*}

\definecolor{Gray}{gray}{0.93}
\begin{table*}
\small
\renewcommand{\arraystretch}{1.4}
\rowcolors{1}{Gray}{Gray}
\resizebox{\textwidth}{!}{
\begin{tabular}{!{\vrule width 1.2pt} p{\linewidth} !{\vrule width 1.2pt}}
    \Xhline{1.2pt}
        Generate a Q\&A pair based on a given context, where the context is understood but NOT DIRECTLY VISIBLE to the person answering the question. Assume the person answering the question has common sense and is aware of the details and key points in the sentence(s), but the sentence(s) itself is not quoted or referenced directly.\\
        \\
        Sentence(s):  \{\textit{sentences}\}\\
        \\
        Use the following instructions for generating a Q\&A pair:\\
        1) Provide a question, and an answer.\\
        2) DON’T use phrases such as ‘according to the sentence(s)’ in your question.\\
        3) An answer should be an entity or entities. Provide a SHORT ANSWER.\\
        4) Generate a SINGLE Q\&A pair.\\
        \\
        Be sure to follow the following format and write your answer within curly brackets.\\
        The format is as follows:\\
        \{Question\}\{Answer\}\\
    \Xhline{1.2pt}
\end{tabular}
}
\caption{Sample prompt for generation of \textproc{New} QA pair (1)}
\label{tab:newQAPrompt}
\end{table*}

\definecolor{Gray}{gray}{0.93}
\begin{table*}
\small
\renewcommand{\arraystretch}{1.4}
\rowcolors{1}{Gray}{Gray}
\resizebox{\textwidth}{!}{
\begin{tabular}{!{\vrule width 1.2pt} p{\linewidth} !{\vrule width 1.2pt}}
    \Xhline{1.2pt}
        Generate a Q\&A pair based on New Sentence(s), where the context in understood but NOT DIRECTLY VISIBLE to the person answering the question. You can reference the Source Content for broader context, but the Q\&A pair should relate directly to the information in New Sentence(s).\\
        \\
        New Sentence(s): \{\textit{sentences}\}\\
        \\
        Source Content : \{\textit{source content}\}\\
        \\
        Use the following instructions for generating a Q\&A pair:\\
        1) Provide a question, and an answer.\\
        2) DON’T use phrases such as ‘according to the sentence(s)’ in your question.\\
        3) An answer should be an entity or entities. Provide a SHORT ANSWER.\\
        4) Generate a SINGLE Q\&A pair.\\
        \\
        Be sure to follow the following format and write your answer within curly brackets.\\
        The format is as follows:\\
        \{Question\}\{Answer\}\\
    \Xhline{1.2pt}
\end{tabular}
}
\caption{Sample prompt for generation of \textproc{New} QA pair (2)}
\label{tab:newQAPrompt2}
\end{table*}

\definecolor{Gray}{gray}{0.93}
\begin{table*}
\small
\renewcommand{\arraystretch}{1.4}
\rowcolors{1}{Gray}{Gray}
\resizebox{\textwidth}{!}{
\begin{tabular}{!{\vrule width 1.2pt} p{\linewidth} !{\vrule width 1.2pt}}
    \Xhline{1.2pt}
        Identify the contradiction between two following sentences and generate a Q\&A pair that reflects this contradiction. The question should be answerable based on each sentence(s), but the two answers should CONTRADICT EACH OTHER. You can reference the Source Content for broader context, but the Q\&A pair should relate directly to the information in Old/New Sentence(s).\\
        \\
        Old Sentence(s) : \{\textit{old sentence}\}\\
        New Sentence(s) : \{\textit{new sentence}\}\\
        \\
        Source Content : \{\textit{source content}\}\\
        \\
        Use the following instructions for generating a Q\&A pair:\\
        1) The question should be answerable based on each sentence.\\
        2) DON’T use phrases such as ‘according to the sentence(s)’ in your question.\\
        3) An answer should be an entity or entities. Provide a SHORT ANSWER.\\
        4) Create a SINGLE Q\&A pair, providing two CONTRADICTORY answers: one based on the old sentence, and another based on the new sentence.\\
        \\
        Be sure to follow the following format and write your answer within curly brackets.\\
        The format is as follows:\\
        \{Question\}\{Answer based on Old Sentence\}\{Answer based on New Sentence\}\\
    \Xhline{1.2pt}
\end{tabular}
}
\caption{Sample prompt for generation of \textproc{Changed} QA pair}
\label{tab:changedQAPrompt}
\end{table*}

\definecolor{Gray}{gray}{0.93}
\begin{table*}
\small
\renewcommand{\arraystretch}{1.4}
\rowcolors{1}{Gray}{Gray}
\resizebox{\textwidth}{!}{
\begin{tabular}{!{\vrule width 1.2pt} p{\linewidth} !{\vrule width 1.2pt}}
    \Xhline{1.2pt}
        Create an Information Dialogue Dataset about \textit{\{topic\}} between two conversation partners (User, Expert).\\ 
        A paragraph about \textit{\{topic\}} will be provided as factual information. The expert's words must be generated to provide an answer based on this information.\\
        \\
        Using the following instruction for generating a dialogue:\\
        1) The user starts the dialogue first\\
        2) Create a multi-turn dialogue of 3-4 turns, each consisting of a not too long conversation.\\
        3) Create it to include each element of conversation, discussion, and QA. In other words, users should not always ask questions using interrogative sentences.\\
        4) DON'T use phrases such as according to the paragraph in guide's utterance.\\
        5) DON'T simply parrot this paragraph or referenced directly. There is no need to include everything given in the paragraph in the dialogue.\\
        6) Do not use what you already know about \textit{\{topic\}}, and the Expert will answer only with the content of the provided paragraph.\\
        7) I will provide you with sentences and a unique number for each sentence. You must indicate the Sentence number you've referenced for each turn.\\
        \\
        Below is an example of output format and dialogues:\\
        \{\{Reference Sentence\}\}2\{\{User\}\}I really love Granny Smith apples, they’re my favorite type of apple\{\{Expert\}\}I love granny smith apples. they have hard, light green skin and a crisp flesh.\\
        \{\{Reference Sentence\}\}1\{\{User\}\}Yes, I really enjoy them. I also like Honeycrisp apples but they’re so expensive!\{\{Expert\}\}they’ve been grown for thousands of years in asia and europe, and were brought to north america by european colonists\\
        \{\{Reference Sentence\}\}3\{\{User\}\}Oh really? They’ve been around way longer than I thought!\{\{Expert\}\}they’re also consumed raw, it’s one of the most popular cooking apples.\\
        \\
        Sentences:\\
        \textit{\{sentences\}}\\
        \\
        Please generate dialogue:\\
    \Xhline{1.2pt}
\end{tabular}
}
\caption{Sample prompt for Dialogue generation}
\label{tab:generateDialoguePrompt}
\end{table*}

\end{document}